%% file: main.tex
\documentclass[10pt,twocolumn,letterpaper]{article}

\usepackage[pagenumbers]{cvpr} %

\input{preamble}

\definecolor{linkblue}{rgb}{0.21,0.49,0.74}
\usepackage[pagebackref,breaklinks,colorlinks,allcolors=linkblue]{hyperref}

\title{One state, many questions: \method gives a frontier model typed eyes for pathology}

\author{Zhixuan Ge \qquad Wei Qiu\textsuperscript{\dag}\\
Rice University\\
{\tt\small \{zg33, wq8\}@rice.edu}\\
{\small \textsuperscript{\dag}Corresponding author}}

\begin{document}
\maketitle
\input{sec/0_abstract}
\input{sec/1_intro}
\input{sec/1b_related}
\input{sec/2_method}
\input{sec/3_experiments}
\input{sec/4_conclusion}
\input{sec/6_acknowledgements}
\input{sec/5_llm_usage}
{
    \small
    \bibliographystyle{ieeenat_fullname}
    \bibliography{main}
}

\input{sec/X_suppl}

\end{document}

%% file: preamble.tex
\usepackage{booktabs}
\usepackage{xspace}
\newcommand{\method}{Opsis\xspace}
\newcommand{\methS}{Opsis-S\xspace}
\newcommand{\methV}{Opsis-V\xspace}

\usepackage{iftex}
\ifXeTeX
  \usepackage[OT1]{fontenc}
\fi

%% file: sec/0_abstract.tex
\begin{abstract}
A synoptic pathology report is a checklist of questions with fixed answer sets, many composed from others. Pathology foundation models answer them accurately with one classifier trained with labels for each, and frontier language models, which compose freely, cannot fit a slide of fourteen thousand tiles in their context. On 189 breast report questions, frontier models that read a written description of the slide or look at a thumbnail and tiles of it answer at 0.50 to 0.66, below the majority class at 0.67. A description is written before the question is known, and a look covers a handful of tiles. We present \method, which lets the model ask instead: a typed question interface on a frozen pathology encoder. A slide is encoded once, a question names an operator, arguments and candidates given at runtime, and the answer is a probability over the candidates. We build it as a 10 M parameter decoder on frozen Virchow2 tokens (\methS) and inside Qwen3-VL-8B (\methV). Asking \methS, four frontier models answer the breast questions at 0.78 to 0.79, and on 3,797 pan cancer questions asking is 8.6 points above reading. On 995 composite questions that \methS was never trained on, a frontier model composing its answers reaches 0.76, against 0.69 when it looks at the slide and 0.78 for ABMIL trained on each composite. On a pan cancer question set over 10,712 TCGA slides, one \methS decoder matches linear probes trained per question on TITAN and PRISM2 embeddings (average accuracy 0.69 against 0.70 and 0.68). \methS also separates CAMELYON16 metastases without training on that cohort (AUROC 0.86) and localises necrosis from report labels alone (tile AUROC 0.93).
\end{abstract}

%% file: sec/1_intro.tex
\section{Introduction}
\label{sec:intro}
\input{fig/fig1_teaser}

A synoptic pathology report is a checklist~\cite{cap_protocols}: histologic type, grade, lymphovascular invasion, necrosis, an in situ component, each a typed question with a small set of allowed answers, and many entries are composed from others, such as a grade from its components or a high grade tumour with lymphovascular invasion. Two kinds of model could fill it. A pathology foundation model encodes a slide into features, and a linear probe or a multiple instance classifier on them answers one question per labelled training run~\cite{zimmermann2024virchow2,chen2024uni,ilse2018abmil}. Without that training, the slide encoders that can be prompted with text or asked a question directly~\cite{ding2024titan,shaikovski2025prism2} are above the majority class on only three and five of eight report fields. With it, a composed question needs labels and a classifier of its own. A frontier language model plans, decomposes and composes, and calls tools when it is given them~\cite{schick2023toolformer,yao2023react}, but it cannot take in a slide of fourteen thousand tiles, the median of our TCGA slides at 20x.

A frontier model can still learn about a slide in three ways: read a written description of it, look at part of it, or ask about it. \Cref{fig:teaser} measures the first two on 189 synoptic and composite questions about breast cancer slides. Reading the 78 word report that PRISM2 generates, or a description of 6 or 48 tiles written with no question in view (927 and 6,140 words), Claude Opus 5.5 and gpt-6-sol answer at 0.50 to 0.63. Looking at a thumbnail and six tiles, four frontier models from two vendors answer at 0.58 to 0.66, and no better when they choose their own views (0.59 and 0.61). None is above the majority class of 0.67. A single tile is not the obstacle: Claude Opus 5.5 and gpt-6-sol separate necrotic from other BCSS tiles at AUROC 0.97 and 0.91. Reading lacks the question and looking lacks coverage. A description is written before the question is known, and at the 128 words per tile of the longer description a whole slide would take 1.8 million words. We therefore let the model ask, and test whether one model on a frozen encoder can answer every entry of the checklist as a probability, at the accuracy of the classifiers trained per entry and cheaply enough to be called as a tool, so that composed entries need no training.

\method, the Greek word for sight, is such a model and the interface it obeys. A frozen pathology encoder gives a field of view or a whole slide one state, encoded once and cached. A question is typed: an operator, its arguments, optional boxes, and a candidate set given at runtime. Questions are isolated, so a batch of them costs one pass over the state and returns the answers of serial execution. Every answer is a probability over the candidates, trained with cross entropy. Composition happens outside the answering model, by a fixed rule or by a frontier model that asks the questions it needs. These four properties define the \method interface.

We instantiate the interface in two ways with complementary strengths. \methS{} is a question decoder of 10\,M parameters, trained once per question set, on the frozen Virchow2 tokens of a field with its context or on sampled tiles of a slide, with per tile heads for the findings that live in a few tiles. Because it reads cached features, it covers whole slides and answers at 0.1\,ms per question once a field is encoded (\cref{sec:cost}), which makes it the tool of the organiser. \methV{} is the same interface on field questions inside Qwen3-VL-8B, which reads the field and its context as pixels, with the two images as a shared prefix and each question as an isolated suffix. \methS{} is more accurate on cell classes and counts, \methV{} on nuclear size and tissue context, and their fusion is the most accurate and best calibrated reading of a field. Above \methS{} sits the organiser, an unmodified frontier model that calls \methS{} as its tools.

Three findings carry the paper. First, one decoder is as accurate as classifiers trained per question. On 1,393 pan cancer test slides the average accuracy of \methS over the 12 slide level fields is 0.69, against 0.70 and 0.68 for the linear probes of TITAN and PRISM2 trained per field on our labels, and on field questions the two instantiations together are 2 to 32 points above a linear probe on the same tokens on five of six question types. Second, asking beats reading and looking, and typed probabilities compose without training. On the 189 breast questions four frontier models answer at 0.78 to 0.79 asking \methS, 14 to 29 points above every reading and looking condition, and on 3,797 questions of the same kinds in the pan cancer test split gpt-6-sol asking is 8.6 points above the same model reading the PRISM2 report (0.72 against 0.64). On the 995 composite questions of that split, which \methS was never trained on, Claude Opus 5.5 composing its answers reaches 0.76, against 0.69 when it looks at the slide and 0.78 for ABMIL~\cite{ilse2018abmil} trained on each composite with labels (\cref{tab:main}). Third, the probabilities do more than score. They localise necrosis from report labels alone (tile AUROC 0.93 against BCSS masks), separate CAMELYON16 metastases~\cite{bejnordi2017camelyon} from negatives at 0.86 without a slide of that cohort in training, and lift the accuracy of \methS on field questions from 0.79 to 0.88 when it abstains on the least confident fifth.

Our contributions are (1) a typed question interface on a frozen pathology state, with a decoder that answers field and slide questions at the accuracy of classifiers trained per question, (2) a controlled study of reading, looking and asking, with composite questions held out of training, (3) the evidence that per operator probabilities localise, transfer and support abstention, and (4) a pan cancer typed question set over 10,712 TCGA slides with audited report labels.

%% file: fig/fig1_teaser.tex
\begin{figure}[t]
\centering
\includegraphics{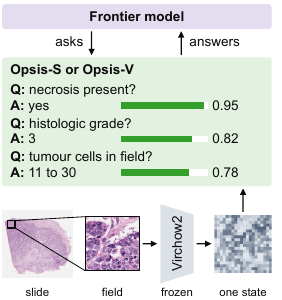}\hfill
\includegraphics{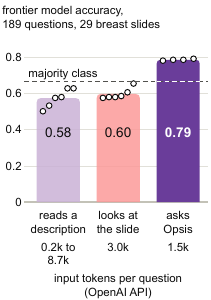}
\caption{\textbf{\method lets a frontier model ask about a slide.} Left: a frozen encoder turns a slide or a field into one state, \methS or \methV answers typed questions about it as probabilities, and an unmodified frontier model decides what to ask and composes the answers. The answers shown are those of \methS on one test slide and the state is the first principal component of the Virchow2 tokens of the field. Right: three ways for a frontier model to take in a slide, on 189 synoptic and composite questions about 29 TCGA-BRCA test slides. Each dot is one model and condition, the bar their mean, the dashed line the majority class (0.67). Reading a written description (the report PRISM2 generates, or a description of 6 or 48 tiles by Claude Opus 5.5) gives 0.50 to 0.63, looking at a thumbnail and six tiles or at views of the model's own choice gives 0.58 to 0.66, and asking \method gives 0.78 to 0.79 (\cref{tab:ways}).}
\label{fig:teaser}
\end{figure}

%% file: sec/1b_related.tex
\section{Related work}
\label{sec:related}

\paragraph{Pathology foundation models.}
Pathology encoders, self supervised on tiles (Virchow2~\cite{zimmermann2024virchow2}, UNI~\cite{chen2024uni}), trained with captions (CONCH~\cite{lu2024conch}) or extended to the whole slide (Prov-GigaPath~\cite{xu2024gigapath}), are read out by linear probes or by multiple instance learning over a bag of tiles~\cite{ilse2018abmil,lu2021clam,shao2021transmil}. Slide encoders such as TITAN~\cite{ding2024titan} and PRISM~\cite{shaikovski2024prism,shaikovski2025prism2} add text prompted zero shot classification and report generation, and PRISM2 also answers yes or no and multiple choice questions with probabilities. On our eight report fields these label free interfaces are above the majority class on three and five (\cref{sec:levels}), so the accurate route remains one trained head per task. \method keeps the frozen encoder and trains one decoder for every typed question.

\paragraph{Pathology vision language models.}
PathChat~\cite{lu2024pathchat}, Quilt-LLaVA~\cite{seyfioglu2024quiltllava} and Patho-R1~\cite{zhang2026pathor1} answer free text questions about regions, SlideChat~\cite{chen2025slidechat} about slides and CPath-Omni~\cite{sun2025cpathomni} about both. Controlled probes find that much of this accuracy survives without the image~\cite{zhang2026fourprobes,zhang2026pathbind}, which is why our question sets are built from annotations and audited report fields and the open vision language models among our comparables are also run with a blank image. A language model can also be given the image as text: captions or a generated report stand in for the pixels~\cite{yang2022pica,zeng2022socratic,guo2023img2llm}, slide level report generators supply such text in pathology~\cite{chen2023wsicaption,guo2024histgen,shaikovski2025prism2}, and ChatCaptioner~\cite{zhu2023chatcaptioner} lets the language model put free text questions to a captioner. A description is fixed before the question is known, and reading one is a control of our study. \method gives the language model typed questions to ask, answered as probabilities.

\paragraph{Tool use and composition.}
Language models call tools~\cite{schick2023toolformer,yao2023react}, compose visual programs from modules~\cite{suris2023vipergpt,gupta2023visprog} and plan over libraries of models~\cite{shen2023hugginggpt}, in oncology with histopathology classifiers among the tools~\cite{ferber2025agent}. The organiser is this pattern with the modules replaced by typed questions to one state, exposed through function calling and the Model Context Protocol~\cite{mcp2024}. Pathology agents navigate the slide itself: PathFinder~\cite{ghezloo2025pathfinder} and SlideSeek~\cite{weishaupt2025slideseek} coordinate several agents, CPathAgent~\cite{sun2025cpathagent} trains one model to navigate, and GIANT~\cite{buckley2025giant} lets a general purpose multimodal model select crops without training. These agents view image regions, the organiser asks typed questions of a cached state, and free viewing is one of our controls. Visual Jev~\cite{yu2026visualjev} reads candidate probabilities for a batch of isolated questions over a shared prefix of a vision language model, a recipe that later work studies further~\cite{zhou2026pixeljev,tommasino2026pathjev,zhang2026pathojev}, and OpenJev-Vision~\cite{openjev2026} trains small readouts on frozen DINOv2 features~\cite{oquab2023dinov2}. \method brings this recipe to a frozen pathology state with box arguments, slide states, per tile heads and a frontier organiser.

%% file: sec/2_method.tex
\section{\method}
\label{sec:method}
\input{fig/fig2_method}

\subsection{The \method interface}
\label{sec:contract}
Let $E$ be a frozen pathology encoder and $x$ a field or a slide. The state $s = E(x)$ is computed once and cached. A question $q = (o, a, b, C)$ has an operator $o$ (count bin, present, grade, and so on), arguments $a$ (a cell class, a tissue class, a region), zero, one or two boxes $b$ in the coordinates of the state, and a candidate set $C$ given at runtime from a fixed candidate vocabulary: a class list, yes or no, or ordered bins. An instantiation $f$ returns $p(c \mid s, q)$ for every $c \in C$. Isolation means that $f$ applied to a batch $\{q_j\}$ returns the same $p(\cdot \mid s, q_j)$ as $f$ applied to $q_j$ alone, so a batch of questions costs one pass over the state and no question can change another. Composition is any function of the answers that is not $f$: a rule, a planner, or a frontier model.

\subsection{One state}
\label{sec:state}
\paragraph{Fields.} A field is resampled to 0.25 mpp and placed on a 448 px canvas, its 448 um surroundings form a second canvas at 1.0 mpp, and the frozen Virchow2 ViT-H/14~\cite{zimmermann2024virchow2} turns each canvas into 32 by 32 tokens of 1,280 dimensions (\cref{fig:method}a). Encoding both canvases takes 42 ms on one H200 and happens once per field.
\paragraph{Slides.} Every 20x tile of 256 px is encoded once into its class token and mean patch token (2,560 dimensions), a median of 14,185 tiles per slide, cached for all 10,712 slides.

\subsection{\methS: a question decoder on the state}
\label{sec:opsis_s}
A state adapter projects the tokens to $d = 384$, adds a source embedding (field or context) and a Fourier embedding of the token position, and passes them through two transformer layers. Each question becomes a short token sequence (\cref{fig:method}b): the operator, the boxes through Fourier features, two box pooled query tokens (the mean of the adapted state tokens whose centre lies inside each box), the arguments and the candidates, each with a slot embedding. A two layer transformer decoder self attends inside each question, with a block diagonal mask so that questions never attend to each other, and cross attends to the state. A linear head on each candidate token gives one logit, and the answer is the softmax over the candidates. Up to 32 questions share one state in one pass. The loss is the cross entropy of the answer plus, on fields, the Poisson likelihood of the per token nucleus counts of the four super classes inside the annotated rectangle, a density auxiliary with weight 1. The decoder has 10 M parameters and trains in 12 minutes per seed on one H200 (\cref{sec:supp_impl}).

\paragraph{Slides.} A decoder of the same architecture, trained on the slide questions, reads a slide through $K = 4{,}096$ tiles sampled from the cached bag at every step and at test time, with the tile position on the slide as its position embedding. For the findings that live in a few tiles (lymphovascular invasion, perineural invasion, necrosis, a precursor component) a per tile head, a two layer MLP on the frozen tile feature, scores every tile of the full bag, and the slide probability is the noisy-OR $1 - \prod_i (1 - p_i)$~\cite{maron1998mil} or the log mean exp of the tile scores. Trained from the slide label alone, the per tile logit is a map of the finding (\cref{sec:transfer}).

\subsection{\methV: the same interface inside a vision language model}
\label{sec:opsis_v}
Qwen3-VL-8B~\cite{qwen3vl2025} receives the field at its native 0.2 mpp, the context canvas and a question with lettered candidates, and the answer distribution is the softmax over the letter logits of the language model head at the answer position (\cref{fig:method}c), the Visual Jev recipe~\cite{yu2026visualjev}. LoRA adapters~\cite{hu2022lora} of rank 16 on the language tower are trained with the cross entropy of the answer letter (2 hours per seed) and the vision tower stays frozen. At inference the two images are a shared prefix whose key value cache is computed once and the questions are isolated suffixes run as one batch, so 32 questions cost 255 ms instead of 2,173 ms, with argmax agreement 0.99 with serial execution.

\subsection{Fusion and routing}
\label{sec:fusion}
When both instantiations run, the answer is the normalised geometric mean of the two probability vectors (\cref{fig:method}d). When latency matters, a router sends a question to \methV by operator, when the confidence of \methS is below a threshold, or by a learned rule (\cref{tab:router}).

\subsection{The organiser}
\label{sec:organiser}
The organiser is a frontier language model that answers a slide level question by calling \methS (\cref{fig:method}e). Five tools are exposed through function calling or the Model Context Protocol: slide information, the slide questions (12 operators, probabilities over the candidates), tile evidence (the slide probability and top tiles of a per tile head), region questions about an 80 um field at a given location (the field decoder, 9 operators with cell and tissue arguments) and, in the conditions where the model also looks at the slide, a view of any region. The prompt, whose example of a composition is high grade with lymphovascular invasion, is in \cref{sec:supp_organiser}. Nothing in the organiser is trained.

%% file: fig/fig2_method.tex
\begin{figure*}[t]
\centering
\includegraphics{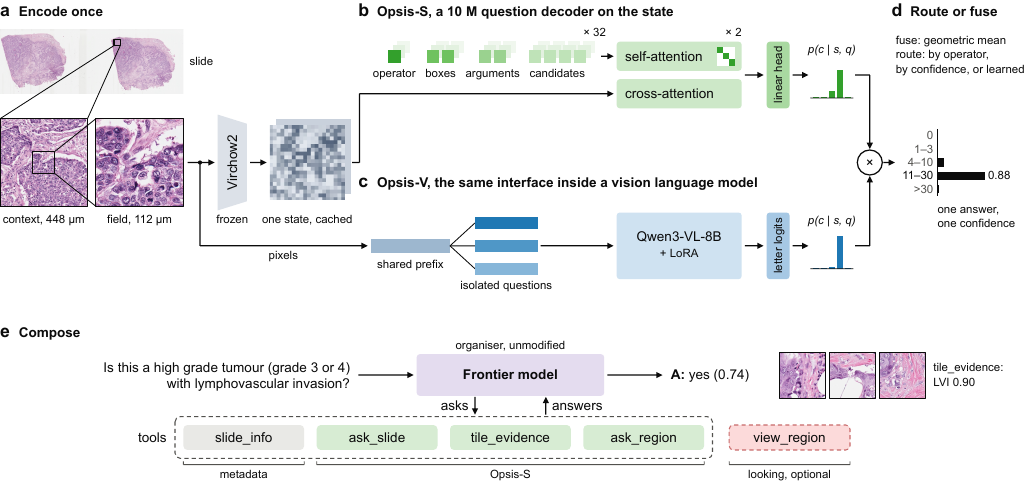}
\caption{\textbf{Overview of \method.} (a) A field and its context are encoded once by the frozen Virchow2 into a cached state. (b) \methS: up to 32 typed questions (operator, boxes, arguments, candidates) pass two decoder layers, with self attention masked between questions and cross attention to the state. (c) \methV: Qwen3-VL-8B with LoRA reads the field and its context as pixels and answers isolated questions on a shared prefix. (d) Fusion by the geometric mean, or routing. The distributions in (b) to (d) are the outputs for this field (tumour cell count). (e) A recorded run of gpt-6-sol on another test slide: the slide question leaves lymphovascular invasion open (0.49), tile evidence supports it (0.90), and the answer is yes, as in the report, where the fixed rule says no. Tools are coloured by what answers them. The tiles are the three the tile head scores highest, unconfirmed by a pathologist. The model receives their coordinates and scores, not pixels.}
\label{fig:method}
\end{figure*}

%% file: sec/3_experiments.tex
\section{Experiments}
\label{sec:exp}
\input{fig/fig3_slide_level}
\input{fig/fig5_transfer_merged}
\input{tables/T0_main_method}

\subsection{Question sets and protocol}
\label{sec:setup}
\paragraph{Fields.} NuCLS~\cite{amgad2022nucls} gives 1,744 fields of 124 TCGA-BRCA patients with nucleus classes, and BCSS~\cite{amgad2019bcss} the tissue classes of their context. From these we build 19 operators in three families, cell questions about one nucleus, region questions about a box and compositional questions about two regions or a compartment, 242,604 questions split by patient (87 train, 19 validation, 18 test). PanNuke~\cite{gamper2019pannuke,gamper2020pannuke} adds 7,901 fields of 19 organs with 9 operators (781,295 questions), trained jointly with NuCLS. The operators are listed in \cref{sec:supp_extended}.

\paragraph{Slides.} The pan cancer set covers 10,712 diagnostic slides of 8,861 patients over 32 TCGA projects~\cite{weinstein2013tcga}. Eight report fields shared across cancers (grade, lymphovascular invasion, perineural invasion, necrosis, precursor or in situ component, mitotic activity, multifocality, lymphocytic infiltrate) were extracted from 8,138 reports of the TCGA-Reports corpus~\cite{kefeli2024tcgareports} by gpt-6-luna~\cite{openai2026models} with a schema of verbatim quotes and audited against an independent extraction (kappa 0.78 to 0.98, \cref{sec:supp_labels}). Cancer type (32 way), molecular subtype, tumour purity~\cite{carter2012absolute} and leukocyte fraction~\cite{thorsson2018immune} come from the clinical tables. Five composite operators are built by rule from report fields and held out of training for every \method model. Slides are split per project 70/15/15 by patient (1,393 test slides) and four projects are held out entirely. The breast set adds the synoptic report of 1,122 TCGA-BRCA slides (18 fields, 170 test slides). Every report derived number is agreement with the report.

\paragraph{Organiser set.} The controlled study asks 189 questions over six synoptic and five composite operators about 29 BRCA test slides (none of their reports states perineural invasion), and 75, 79 and 75 questions about LUAD, BLCA and UCEC slides. Each condition answers the same questions: \methS with fixed rules and no language model (four composites have a rule, precursor type has none and counts as wrong), and a frontier model looking at the slide (a thumbnail and six 256 um tissue tiles) or with \method tools. Four frontier models answer with the same tools: gpt-6-sol, gpt-5.5 and gpt-6-astra~\cite{openai2026models} and Claude Opus 5.5~\cite{anthropic2026opus} (\cref{sec:supp_organiser}). Two controls take the place of asking: free viewing, where the tools return only pixels of any region the model chooses, and a written description, where each question is answered from a text written with no question in view, the report that PRISM2 generates or a description of 6 or 48 tiles by Claude Opus 5.5. Claude Opus 5.5 with the tools also answers the 995 composite questions of the pan cancer test split, which fill the composite rows of \cref{tab:main}, and gpt-6-sol the 3,797 questions of seven report fields and the five composites. These runs name the slide by a hash and the tools return no project, whereas the controlled study names each slide by its TCGA barcode in every condition with tools or images except free viewing by Claude Opus 5.5 (\cref{sec:supp_organiser}).

\paragraph{Comparables and metrics.} The comparables trained per task with labels serve as references: CellViT++ with the Virchow backbone~\cite{horst2024cellvit,horst2025cellvitpp} for cells and ABMIL~\cite{ilse2018abmil} for slides, one classifier per field on the same frozen Virchow2 tile features \methS reads. Those without labels for the task are TITAN and PRISM2 prompted with text, Qwen3-VL-8B and Patho-R1-7B~\cite{zhang2026pathor1} answering the typed questions from the same images as \methV, and Claude Opus 5.5 looking at the same images. CLAM-SB~\cite{lu2021clam}, linear probes on Virchow2 and the linear probes of TITAN and PRISM2, which align slide embeddings with pathology reports in pretraining, are in the appendix (\cref{tab:fov,tab:slide,tab:pan}). We report accuracy against the majority class, negative log likelihood and expected calibration error~\cite{guo2017calibration}, as the mean over three seeds for the \method models.

\subsection{One interface across levels}
\label{sec:levels}
\Cref{tab:main} compares \method with the standard method of each level. Overall, \method is the most accurate method not trained per task on 15 of the 22 accuracy rows. It is above the per task reference on 8 of 18 rows and within five points on 15. The composite rows are taken up in \cref{sec:slide}.

\paragraph{\method vs.\ field level comparables.} On the field questions the fusion of \methS and \methV outperforms every comparable of \cref{tab:main} on all six rows. It reaches 0.816 on the nucleus super class, above CellViT++ trained on the same nuclei with their locations given (0.804), 0.748 on the seven nucleus classes, above CellViT++ with a head trained on them (0.727), and 0.805 over all 28,735 test questions (\cref{tab:fov}) with the best calibration (ECE 0.026). The two instantiations are complementary: \methS leads on classes and counts, \methV on nuclear size and tissue context, and an oracle choice between them reaches 0.861. A linear probe on the same frozen Virchow2 tokens, trained per task with labels, is 2 to 32 points below the fusion on five of the six rows and leads on the stromal TIL score (0.717 against 0.663 on 99 questions). Qwen3-VL-8B and Patho-R1-7B answering the same typed questions reach 0.421 and 0.544 over all test questions with the images and 0.494 and 0.513 with a blank image, so their answers come from the question more than from the tissue. Claude Opus 5.5 is 3 to 40 points below the fusion.

\paragraph{\method vs.\ slide encoders prompted with text.} On the pan cancer fields \methS is above the majority class on every field, whereas TITAN and PRISM2 prompted with text are above it on three and five of the eight report fields (\cref{fig:slide}a). \methS outperforms both on the 32 way cancer type (0.937 against 0.877 and 0.707) and, by 3 to 19 points, on necrosis, lymphovascular invasion and tumour purity. PRISM2 leads on perineural invasion (0.790 against 0.771) and, on the breast report, on grade and lymphovascular invasion by 1 and 3 points (\cref{tab:slide}). Claude Opus 5.5 looking at the slide is on par with \methS on mitotic activity (0.779 against 0.771) and at most seven points above the majority class on necrosis, lymphovascular invasion and tumour purity.

\paragraph{\method vs.\ supervised MIL.} \methS is on par with the reference, ABMIL trained per field on the same tiles, within one point on every pan cancer field of the sample, each as the mean over three seeds. ABMIL leads on lymphovascular invasion of the breast report (0.730 against 0.694).

\paragraph{\methS vs.\ probes trained per field.} Averaged over all 12 pan cancer fields, the one \methS decoder (0.686) is on par with the linear probes of TITAN (0.697) and PRISM2 (0.683), one classifier per field trained on our labels, and with a linear probe on the mean Virchow2 vector of the slide (0.689, \cref{tab:pan}). On 10 of the 12 fields the probes, ABMIL and \methS with its tile head are within five points of each other. The exceptions are mitotic activity, where the TITAN probe leads by nine points, and the lymphocytic infiltrate on 40 questions, where \methS leads. On the four held out projects the probes and \methS lose accuracy together: \methS stays above the majority class on grade, lymphovascular invasion and necrosis and falls below it on the other five report fields, and PRISM2 prompted with text is above it on five (\cref{tab:pan_heldout}). What differs is the interface: a probe is one labelled training run per field, \methS is one decoder for every field.

\subsection{Reading, looking and asking}
\label{sec:slide}

\paragraph{Asking vs.\ looking and reading.} \Cref{fig:slide}b and \cref{tab:ways} give the controlled study on 29 breast slides and 189 questions. Overall, a frontier model asking \method is 12 to 13 points above the majority class and 14 to 29 points above the same model looking at the slide or reading a description of it. gpt-6-sol answers at 0.788 asking \methS and at 0.582 looking at the slide, and the gap is largest on the fields that need the whole slide, necrosis (0.846 against 0.385) and lymphovascular invasion (0.810 against 0.381). gpt-5.5, gpt-6-astra and Claude Opus 5.5 reproduce the picture (0.783, 0.788 and 0.794 asking, 0.582, 0.577 and 0.656 looking, against a majority class of 0.67). Looking longer does not close the gap: choosing their own regions, gpt-6-sol and Claude Opus 5.5 reach 0.587 and 0.608. Reading does not close it either: from the 78 word report of PRISM2 or from the descriptions of 6 and 48 tiles (927 and 6,140 words), Claude Opus 5.5 and gpt-6-sol reach 0.50 to 0.63, and the PRISM2 report states grade for 49 percent of the slides, lymphovascular invasion for 61 and necrosis for 31 (\cref{tab:pan}). Every one of these conditions is below the majority class, and the bootstrap interval over slides of its gap to asking excludes zero. The gap holds beyond the 29 slides: on the 3,797 questions about seven report fields and the five composites in the pan cancer test split (1,179 slides), gpt-6-sol answers at 0.722 asking and at 0.637 reading the PRISM2 report, 8.6 points apart (95 percent interval 6.6 to 10.5, \cref{tab:ways_pan}).

\paragraph{Composition vs.\ training on the composite.} Composite questions test whether the answers compose, since no \method model was trained on them. On the 995 composite questions of the pan cancer test split (\cref{tab:main}), Claude Opus 5.5 composing them from the primitive tools reaches 0.757. It outperforms the same model looking at the slide (0.685) and is within three points of ABMIL trained on each composite with labels (0.784). Fixed rules over the \methS probabilities exist for four of the five composites, where they are on par with the organiser (0.78 each), and only the organiser answers precursor type. On the largest composite, high grade with lymphovascular invasion, the organiser outperforms ABMIL (0.797 against 0.781), and it is close on grade from its components (0.673 against 0.694). It trails ABMIL on necrosis with high mitotic activity (0.822 against 0.879), on DCIS with comedo necrosis (0.660 against 0.713 on 50 questions) and on precursor type (0.685 against 0.784), where no tool names the type and the organiser infers it from the precursor and cancer type questions.

In the controlled study on breast slides the rules and the organiser differ by at most one question where a rule exists, and the organiser alone answers precursor type (0.79), which accounts for its lead over the rules (0.788 against 0.704). The rules thus confirm the compositions of the organiser where a rule exists, and the organiser needs no rule to be written, which is what lets it answer precursor type. \Cref{fig:slide}c repeats the conditions on lung, bladder and endometrium, where the TCGA reports state grade and invasion sparsely and \methS with rules answers at 0.533, 0.582 and 0.627 against majority classes of 0.600, 0.747 and 0.587. There asking stays within four points of the rules: the organiser is only as good as the model it asks. 

\subsection{Localisation and transfer}
\label{sec:transfer}
\Cref{fig:transfer}a shows a BCSS region of a held out slide: the necrosis logit of the \methS per tile head, trained from the report field alone, outlines the necrotic region as the annotation does. Over 8,830 tiles of 119 BCSS slides it tracks the masks at AUROC 0.933 (0.927 on the 20 slides in the test split of the tile heads), on par with the attention of the same tile head with gated attention pooling~\cite{ilse2018abmil} trained on the same labels (0.942 and 0.920, \cref{fig:transfer}b). Interestingly, the attention is not specific: the attention of the precursor head also follows necrosis (0.70), whereas the precursor logit of \methS does not (0.23).

On CAMELYON16~\cite{bejnordi2017camelyon}, a cohort, scanner and tissue the field decoder never saw, the probability of tumour cells on 80 um fields separates lesion from lesion free fields at AUROC 0.951 on 1,170 fields of 39 test slides (\cref{fig:transfer}d). Read over every tissue field of the 129 official test slides and aggregated by the mean of the five highest probabilities, a rule chosen on slides of the challenge training set, it separates macro metastases from negatives at 0.992, micro metastases at 0.759 and all positives at 0.864, against 0.966 for the pathologist without time constraint, 0.994 for the best challenge entry trained on the cohort and 0.968 for PRISM2 prompted with text (\cref{fig:transfer}e). The ranking transfers and the threshold does not: half of the lesion fields are answered with zero tumour cells, so a new cohort needs a recalibrated threshold or abstention.

\subsection{Cost, confidence and ablations}
\label{sec:cost}
\methS answers 32 questions in 2.9 ms after the 42 ms of encoding, 32 times faster than serial execution, and \methV with the shared prefix is 8.5 times faster than recomputing it. Abstaining on the 20 percent least confident questions~\cite{geifman2017selective} lifts \methS from 0.791 to 0.876 and the fusion from 0.805 to 0.891. On held out PanNuke organs accuracy drops from 0.842 to 0.783 and the mean confidence of a field flags an unseen organ at AUROC 0.69 (\cref{fig:cost}).

\label{sec:ablations}
\Cref{tab:ablations} in the appendix changes one factor of \methS at a time. Four parts of the decoder matter. Without the box pooled query tokens accuracy drops by 0.021 and the comparison of two regions by 0.29. One fixed linear head per operator and argument pair in place of the question conditioned attention costs 0.015 overall and 0.25 on nuclear size. Training on NuCLS alone instead of jointly with PanNuke costs 0.051 on the cell family, and removing the context canvas costs 0.015. Nothing that enlarges the model moves accuracy by a point: a decoder of 768 dimensions and three plus three layers, 128 instead of 32 questions per step and an 896 px state change it by 0.009 or less, and unfreezing the last four Virchow2 blocks changes the test likelihood by 0.02 (\cref{sec:supp_ablations}). \methV reading only the pooled Virchow2 tokens of the state instead of its own image tokens reaches the same likelihood as the pixel path (0.563 against 0.563), so one state can serve both instantiations.

%% file: fig/fig3_slide_level.tex
\begin{figure*}[t]
\centering
\includegraphics[width=0.95\textwidth]{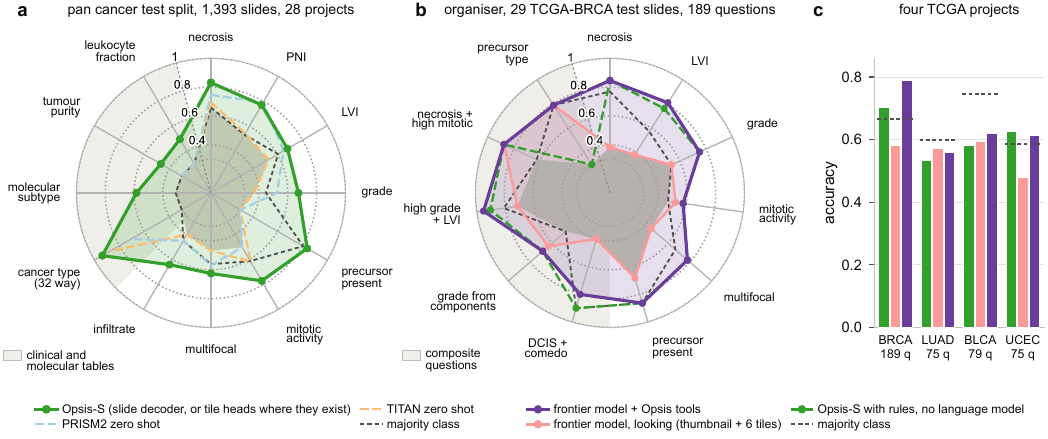}
\caption{\textbf{The slide level.} (a) Pan cancer test split (1,393 slides of 28 projects), one axis per field, the clinical and molecular table fields in the shaded sector, radial scale 0.3 at the innermost ring to 1 at the rim: \methS (the slide decoder, or its full bag tile head on the fields that have one) against TITAN and PRISM2 prompted with text and the majority class. (b) The organiser study on 29 TCGA-BRCA test slides, 189 questions, one axis per operator, the composite questions in the shaded sector: \methS with fixed rules and no language model (zero on precursor type, which has no rule), gpt-6-sol looking at a thumbnail and six tiles, and gpt-6-sol asking \method. (c) The same three conditions on four TCGA projects in the colours of (b), the dashed line is the majority class. In lung, bladder and endometrium asking stays within four points of the rules: the organiser is only as good as the model it asks.}
\label{fig:slide}
\end{figure*}

%% file: fig/fig5_transfer_merged.tex
\begin{figure*}[t]
\centering
\includegraphics[width=0.93\textwidth]{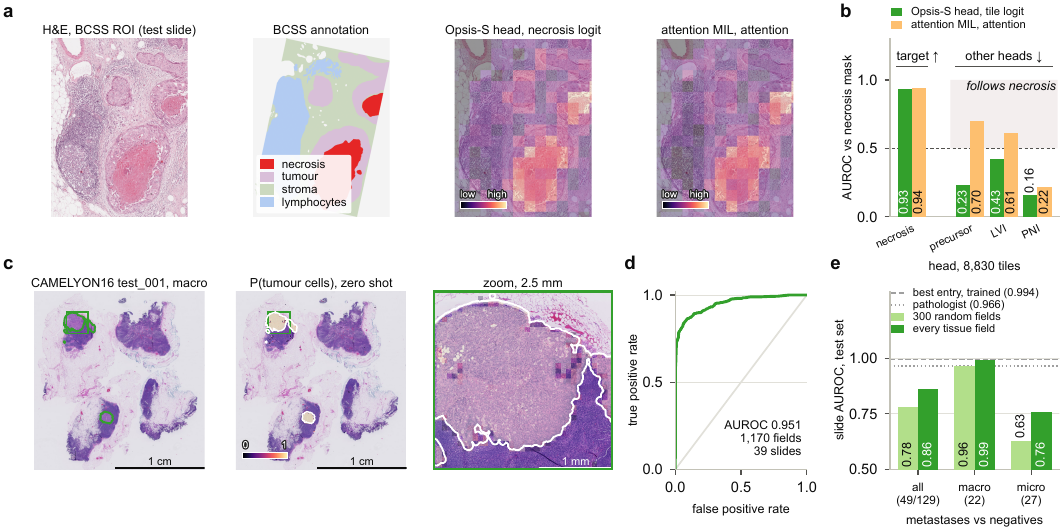}
\caption{\textbf{Localisation and transfer.} (a) A BCSS region of a held out TCGA-BRCA slide: H\&E, the BCSS annotation, the necrosis logit of the \methS tile head trained on report labels alone, and the attention map of the same tile head with gated attention pooling. (b) AUROC of each head's tile score against the necrosis mask (8,830 tiles): the necrosis head should be high, the other heads low. (c) CAMELYON16, never seen in training: a macro metastasis test slide with the reference lesion outlined, the field probability of tumour cells from \methS on an 80 um grid, and a zoom on the lesion. (d) Field level ROC against the lesion masks on 1,170 fields of 39 test slides. (e) Slide level AUROC on the official 129 slide test set for all positives, macro metastases and micro metastases against negatives, from 300 random fields per slide or from every tissue field. Reference lines: best challenge entry (0.994), pathologist without time constraint (0.966).}
\label{fig:transfer}
\end{figure*}

%% file: tables/T0_main_method.tex
\begin{table*}[t]
\centering
\footnotesize\setlength{\tabcolsep}{5.5pt}
\caption{\textbf{One interface, questions at every level.} A sample of tasks against the comparables of each level, the full benchmarks are in \cref{tab:main_components,tab:fov,tab:slide,tab:pan,tab:pan_heldout}. Bold is the best and underline the second best outside the grey columns: the majority class (maj.) and the references trained per task with labels, CellViT++ (the Virchow variant trained on NuCLS) for cells and ABMIL (one classifier per field on the frozen Virchow2 tile features \methS reads, three seeds) for slides. Without labels for the task: TITAN and PRISM2 prompted with text (zs), and Qwen3-VL-8B, Patho-R1-7B and Claude Opus 5.5 answering the typed questions from the field and its context on the field rows and from a thumbnail and six tiles on the slide rows. The \method column is the fusion of \methS and \methV on field questions, \methS on slide questions (its full bag tile head on the fields that have one), and Claude Opus 5.5 composing \methS answers on the 995 composite questions of the pan cancer test split, so the Opus 5.5 and \method columns of these rows are the same model looking and asking, and ABMIL is trained on the composite labels. A cell is blank where the model was not run. The 32 cancer type candidates include the four held out projects.}
\label{tab:main}
\begin{tabular}{lrrrrrrrr}
\toprule
& & \multicolumn{1}{c}{\makebox[0pt]{reference}} & \multicolumn{5}{c}{no labels for the task} & \multicolumn{1}{c}{ours} \\ \cmidrule(lr){3-3} \cmidrule(lr){4-8} \cmidrule(lr){9-9}
task & \rotatebox{90}{\textcolor[gray]{0.45}{maj.}} & \rotatebox{90}{\textcolor[gray]{0.45}{per task}} & \rotatebox{90}{TITAN zs} & \rotatebox{90}{PRISM2 zs} & \rotatebox{90}{Qwen3-VL} & \rotatebox{90}{Patho-R1} & \rotatebox{90}{Opus 5.5} & \rotatebox{90}{Opsis} \\
\midrule
\multicolumn{9}{l}{\textit{Field level, NuCLS test split (99 to 2,399 questions per row), accuracy. \method: fusion of \methS and \methV}} \\
cell super class & \textcolor[gray]{0.45}{0.47} & \textcolor[gray]{0.45}{0.804} &  &  & 0.471 & 0.484 & \underline{0.715} & \textbf{0.816} \\
cell class & \textcolor[gray]{0.45}{0.47} & \textcolor[gray]{0.45}{0.727} &  &  & 0.186 & 0.462 & \underline{0.668} & \textbf{0.748} \\
nuclear size bin & \textcolor[gray]{0.45}{0.35} &  &  &  & 0.309 & 0.176 & \underline{0.432} & \textbf{0.834} \\
compare regions & \textcolor[gray]{0.45}{0.52} &  &  &  & 0.491 & 0.484 & \underline{0.811} & \textbf{0.839} \\
dominant tissue & \textcolor[gray]{0.45}{0.51} &  &  &  & 0.442 & 0.561 & \underline{0.607} & \textbf{0.688} \\
sTIL score bin & \textcolor[gray]{0.45}{0.43} &  &  &  & 0.152 & \underline{0.434} & 0.424 & \textbf{0.663} \\
\addlinespace[2pt]\multicolumn{9}{l}{\textit{Slide level, TCGA-BRCA test split (170 slides), accuracy. \method: \methS}} \\
ER positive & \textcolor[gray]{0.45}{0.72} & \textcolor[gray]{0.45}{0.858} & 0.463 & \underline{0.778} & 0.710 & 0.648 & 0.636 & \textbf{0.813} \\
histologic grade & \textcolor[gray]{0.45}{0.48} & \textcolor[gray]{0.45}{0.674} & 0.483 & \textbf{0.697} & 0.414 & 0.331 & 0.586 & \underline{0.690} \\
LVI & \textcolor[gray]{0.45}{0.70} & \textcolor[gray]{0.45}{0.730} & 0.400 & \textbf{0.727} & 0.300 & \underline{0.700} & 0.673 & 0.694 \\
PAM50 subtype & \textcolor[gray]{0.45}{0.48} & \textcolor[gray]{0.45}{0.671} & 0.235 & \underline{0.524} & 0.229 & 0.482 & 0.459 & \textbf{0.655} \\
\addlinespace[2pt]\multicolumn{9}{l}{\textit{Slide level, pan cancer test split (1,393 slides of 28 projects), accuracy. \method: \methS}} \\
cancer type (32) & \textcolor[gray]{0.45}{0.12} & \textcolor[gray]{0.45}{0.933} & \underline{0.877} & 0.707 & 0.112 & 0.123 & 0.659 & \textbf{0.937} \\
necrosis & \textcolor[gray]{0.45}{0.66} & \textcolor[gray]{0.45}{0.829} & 0.694 & \underline{0.746} & 0.607 & 0.467 & 0.558 & \textbf{0.831} \\
LVI & \textcolor[gray]{0.45}{0.60} & \textcolor[gray]{0.45}{0.686} & 0.528 & \underline{0.649} & 0.407 & 0.594 & 0.615 & \textbf{0.679} \\
PNI & \textcolor[gray]{0.45}{0.53} & \textcolor[gray]{0.45}{0.780} & 0.543 & \textbf{0.790} & 0.554 & 0.473 & 0.710 & \underline{0.771} \\
mitotic activity & \textcolor[gray]{0.45}{0.61} & \textcolor[gray]{0.45}{0.770} & 0.616 & 0.502 & 0.530 & 0.724 & \textbf{0.779} & \underline{0.771} \\
tumour purity & \textcolor[gray]{0.45}{0.27} & \textcolor[gray]{0.45}{0.464} & 0.281 & 0.260 & 0.260 & 0.251 & \underline{0.339} & \textbf{0.468} \\
\addlinespace[2pt]\multicolumn{9}{l}{\textit{Composites held out of training (995 questions on 688 test slides), accuracy. \method: Opus 5.5 asking \methS}} \\
high grade with LVI & \textcolor[gray]{0.45}{0.76} & \textcolor[gray]{0.45}{0.781} & 0.451 & 0.503 & 0.284 & \underline{0.735} & 0.718 & \textbf{0.797} \\
precursor type & \textcolor[gray]{0.45}{0.50} & \textcolor[gray]{0.45}{0.784} & \underline{0.612} & 0.505 & 0.198 & 0.559 & 0.608 & \textbf{0.685} \\
necrosis with high mitotic & \textcolor[gray]{0.45}{0.71} & \textcolor[gray]{0.45}{0.879} & 0.559 & \textbf{0.842} & 0.404 & 0.815 & 0.815 & \underline{0.822} \\
grade from components & \textcolor[gray]{0.45}{0.51} & \textcolor[gray]{0.45}{0.694} & 0.531 & 0.653 & 0.347 & 0.429 & \textbf{0.694} & \underline{0.673} \\
DCIS with comedo necrosis & \textcolor[gray]{0.45}{0.74} & \textcolor[gray]{0.45}{0.713} & 0.620 & 0.340 & \textbf{0.780} & 0.340 & 0.320 & \underline{0.660} \\
all 995 composite questions & \textcolor[gray]{0.45}{0.67} & \textcolor[gray]{0.45}{0.784} & 0.519 & 0.560 & 0.314 & 0.657 & \underline{0.685} & \textbf{0.757} \\
\addlinespace[2pt]\multicolumn{9}{l}{\textit{Transfer without training on the dataset, AUROC. \method: \methS}} \\
CAMELYON16 metastasis, 129 slides &  &  & 0.825 & \textbf{0.968} & 0.484 & 0.511 & 0.680 & \underline{0.864} \\
BCSS necrosis, 8,830 tiles &  &  &  &  & 0.780 & \underline{0.958} & \textbf{0.970} & 0.933 \\
\bottomrule
\end{tabular}
\end{table*}

%% file: sec/4_conclusion.tex
\section{Conclusion}
\label{sec:conclusion}
\method is an interface more than a model: one frozen state, isolated typed questions, probabilities as answers, and composition outside the answering model. Its two instantiations answer field questions, and \methS slide questions, on par with classifiers trained per question, in one pass and with a probability that supports abstention, maps the finding on the slide and transfers to an unseen cohort. A frontier model that asks \methS answers composite questions the decoder was never trained on, and on the breast questions it is 14 to 29 points more accurate than when it reads a description of the slide or looks at it. Asking is bounded by the model that answers: where \methS is below the majority class, as in lung and bladder, so is the organiser. A pathologist's report is a set of typed questions, and letting a reasoner ask them is the direct way to fill it. The code, the question sets and the \methV adapter are available at \url{https://github.com/gezhixuan/opsis}.

%% file: sec/6_acknowledgements.tex
\section*{Acknowledgements}
The results shown here are in whole or part based upon data generated by the TCGA Research Network: \url{https://www.cancer.gov/tcga}.

%% file: sec/5_llm_usage.tex
\section*{LLM Usage Statement}
Claude (Anthropic) assisted with code, figure scripts, literature
search, discussion of ideas and refinement of text written by the
authors. Claude and ChatGPT (OpenAI) gave reviewer-style feedback on
drafts. The pan cancer report fields were extracted from the free text
TCGA reports by gpt-6-luna, with a verbatim quote required for every
value and checked against the source report, and audited against an
independent extraction by Claude Sonnet 5.5 on 320 stratified reports.
The language models evaluated in our experiments are objects of study
and are described in the main text. The authors verified all results
and take full responsibility for the content.

%% file: sec/X_suppl.tex
\clearpage
\appendix
\renewcommand{\thesection}{\Alph{section}}

\noindent The question sets are released as records keyed by TCGA, NuCLS and PanNuke identifiers, with the audited report fields and their verbatim quotes.

\section{Implementation details}
\label{sec:supp_impl}
\paragraph{States.} Field canvases are 448 px at 0.25 mpp (the NuCLS field resampled from 0.2 mpp and centre padded with white, the field occupying a recorded sub window) and 448 px at 1.0 mpp for the 448 um context. PanNuke fields are 256 px at 0.25 mpp without a context canvas. Virchow2 (ViT-H/14, patch 14) gives 32 by 32 patch tokens of 1,280 dimensions per canvas, stored in fp16. Slide tiles are the 256 px tiles at 20x of the TCGA diagnostic slides, each represented by the concatenation of the Virchow2 class token and the mean of its patch tokens (2,560 dimensions).

\paragraph{\methS on fields.} State adapter: linear 1,280 to 384, source embedding (field, context), Fourier position features of the token centre (8 frequencies), two pre norm transformer encoder layers with 8 heads and width 1,536, dropout 0.1. Question tokens: operator embedding, Fourier features of the box and of the second box, two box pooled tokens, up to four argument embeddings and up to 32 candidate embeddings, plus a slot embedding per position. Decoder: two pre norm transformer decoder layers, block diagonal self attention over the question tokens, cross attention to the adapted state. Density head: a linear layer on each field token predicting the log rate of nuclei of the four nucleus super classes (tumour, stromal, sTIL, other) in that token's 14 px cell, trained with the Poisson likelihood inside the annotated rectangle (weight 1). Optimiser AdamW~\cite{loshchilov2019adamw}, learning rate $3 \times 10^{-4}$, weight decay 0.05, batch of 8 states with 32 questions each, 30 epochs with early stopping on validation NLL (patience 6, stopped around epoch 22), NuCLS and PanNuke batches interleaved with 1,800 PanNuke states per epoch. Three seeds.

\paragraph{\methS on slides.} The same adapter and decoder over $K = 4{,}096$ tiles sampled without replacement from the cached bag (tiles repeated when the bag is smaller), with the normalised slide coordinates of each tile as its position, no context canvas, learning rate $2 \times 10^{-4}$, batch of 4 slides, early stopping with patience 8 (3.6 hours per seed, stopped around epoch 13). Composite operators were excluded from training and are evaluated held out. The breast model is the same decoder trained on the breast question set.

\paragraph{Per tile heads.} For lymphovascular invasion, perineural invasion, necrosis and the precursor component, an MLP (layer norm, 2,560 to 512 to 512 to 4 logits, GELU, dropout 0.1) scores every tile of the full bag. Pooling: noisy-OR of the tile probabilities, log mean exp of the tile logits, or gated attention~\cite{ilse2018abmil} for the attention pooled tile head of \cref{sec:transfer}, whose tile score is the attention logit. Output bias initialised at $-10$ with a ten times larger learning rate, AdamW at $10^{-4}$, weight decay 0.01, eight slides per step, 12 epochs with patience 4. Both the noisy-OR head and the log mean exp head are reported as the mean over three seeds.

\paragraph{\methV.} Qwen3-VL-8B-Instruct, LoRA rank 16 (alpha 32, dropout 0.05) on the query, key, value, output, gate, up and down projections of the language tower, vision tower frozen. Prompt: the system line ``You are a pathology image analysis assistant. Answer each multiple choice question about the images with the option letter only.'', the field at its native 0.2 mpp and the context canvas, the question text with the boxes given as normalised coordinates, and the candidates as lettered options. Training: cross entropy of the answer letter token, 3 epochs, 16 questions per state, learning rate $10^{-4}$, bf16 with flash attention, about 2 hours per seed on one H200. Inference with the shared prefix: the key value cache of the image prefix is computed once per state and the question suffixes are run as one right padded batch.

\paragraph{Fusion and routers.} Fusion: elementwise geometric mean of the two probability vectors, renormalised. Routers (\cref{tab:router}): per operator (the instantiation with the better validation accuracy per operator), confidence escalation (questions whose \methS confidence is in the lowest 10, 25 or 50 percent of the validation distribution go to \methV), and a learned router (logistic regression on the operator, the \methS confidence and its entropy, trained on validation to predict which instantiation is correct). Costs assume 41.7 ms of encoding per state, and a question that a confidence or learned router escalates is charged for both instantiations.

\paragraph{Trained models.} \method consists of five sets of weights on one frozen Virchow2 encoder: the field decoder (NuCLS with PanNuke), the pan cancer slide decoder, the breast slide decoder, the per tile heads, and the LoRA adapters of \methV, which by default reads a field and its context canvas as pixels.

\paragraph{Comparables.} CellViT++: the released Virchow backbone with its NuCLS super class token classifier retrained on our training patients, evaluated on ground truth nuclei of the test patients. ABMIL: the gated attention MIL of Ilse et al., the attention network and classifier of the CLAM code without its instance clustering loss, one model per field on the Virchow2 bags of our split (three seeds, full bags on the breast fields and 4,096 random tiles per slide on the pan cancer fields and composites). CLAM-SB: the public implementation with repository defaults on the same bags, one model per field, three seeds on seven breast fields, one seed on the pan cancer fields. Virchow2 linear probe: logistic regression on the mean of the Virchow2 tile vectors of the slide (up to 8,192 random tiles), and on the field questions on the question's box pooled from the cached tokens as the decoder pools it, the second box and the mean field token, one classifier per operator and argument group, C chosen on validation. TITAN probes: logistic regression on the released TCGA slide embeddings (768 dimensions), one per field, regularisation chosen on validation. PRISM2 probes: the same on the PRISM2 slide embeddings computed from our Virchow2 class tokens. TITAN and PRISM2 zero shot: the models' text prompt interfaces with one prompt set per field. PRISM2 report: the ``Write a report'' generation for every test and held out slide, extracted with the pan cancer schema and scored per field with an unstated field counted as wrong. Qwen3-VL-8B and Patho-R1-7B: the same images and lettered questions as \methV, untrained, and once more with a uniform grey image in place of each image as the language prior.

\section{The organiser protocol}
\label{sec:supp_organiser}
Each question is one episode. The model receives a system prompt (``You are a pathologist's assistant answering one multiple choice question about a whole slide image.'', followed in the tool conditions by ``You have tools that read the slide with trained models, they return calibrated probabilities. Compose composite questions from the primitive questions, for example high grade AND lymphovascular invasion from pan\_grade and pan\_lvi, and prefer the tool probabilities over guessing.'' and in the looking conditions by ``You can look at the slide yourself: a thumbnail and tiles are given, and view\_region shows any location.''), the question with lettered options, and in the looking condition a 1,024 px thumbnail and six 256 um tissue tiles at 0.5 mpp drawn with a fixed seed. It answers with a JSON object (option letter, confidence, one sentence) within eight tool steps. Tools: \texttt{slide\_info}, \texttt{ask\_slide} (operators and probabilities of \methS over 4,096 tiles), \texttt{tile\_evidence} (slide probability and top tiles of the log mean exp tile head), \texttt{ask\_region} (the field decoder on an 80 um field at a level 0 location, with cell class and tissue arguments) and, in the looking conditions, \texttt{view\_region} and a thumbnail. The API conditions use the OpenAI Responses API with function calling, the Codex conditions the Codex CLI with the same tools served over the Model Context Protocol in read only mode, where the CLI's cached system prompt adds about 90k input tokens per question. Claude Opus 5.5 runs through the Claude Code CLI in print mode with the same system prompt, the same server restricted to the tools of its condition, and no built in tool except a file read of the prescribed thumbnail and tiles in the looking condition, where the CLI's cached preamble and the image files bring the count to 6.7k input tokens per question with tools and 50k looking. On the pan cancer composite questions the organiser sees each slide under an opaque name that the tool server resolves, and \texttt{slide\_info} withholds the project and the patient, so the TCGA barcode never reaches the model. The rules condition answers the primitive operators with \methS and four of the five composites with a fixed rule over its probabilities, each against a threshold of 0.5: high grade with LVI as the probability of a high grade times that of lymphovascular invasion, necrosis with high mitotic activity as the product of the two, grade from its components as the grade with the largest probability among the candidates, and DCIS with comedo necrosis as the probability of a precursor times that of necrosis. The pan cancer operators include no tubule, pleomorphism or DCIS question, so the last two rules use the nearest primitives. They were added after the first runs of the study. Precursor type has no rule and counts as wrong. On the pan cancer test split the four rules score 0.782, 0.890, 0.680 and 0.640, together 0.78 on 773 questions, as does Claude Opus 5.5 asking (0.7805 and 0.7775). The tool calls of Claude Opus 5.5 on these questions show what it composes. It asks the slide questions on 98 to 100 percent of the questions of every composite: grade and lymphovascular invasion for high grade with LVI (with tile evidence on 26 percent), necrosis and mitotic activity for necrosis with high mitotic activity, grade and mitotic activity for grade from its components, precursor and necrosis for DCIS with comedo necrosis (with tile evidence on 96 percent and region questions on 52 percent, 5.2 calls per question against 1.4 to 2.0 on the others), and precursor and cancer type for the precursor type, which it then names from its own knowledge. Results per condition and model are in \cref{tab:ways}, and per field and composite in \cref{tab:main_components,tab:ways_pan}.

\paragraph{Controls.} Two controls replace asking (\cref{tab:ways}). In free viewing the model has \texttt{slide\_info}, a thumbnail and \texttt{view\_region}, with no trained model behind any tool, and chooses what to look at (gpt-6-sol viewed 10.1 regions per question, Claude Opus 5.5 5.4). In the written description control a slide is described once with no question in view and each question is then answered from that text alone, with no image and no tool. The text is the report PRISM2 generates for the slide (78 words on average), or a description by Claude Opus 5.5 of the thumbnail and the six tiles of the looking condition (927 words, 58k tokens to write), or of these and 42 more random tissue tiles (6,140 words, 351k tokens), one paragraph per image, written for a colleague who cannot see the slide. gpt-6-sol answers from the same cached texts. In the grey image control every image of the looking condition is replaced by a grey image of the same size.

\section{Report labels and their audit}
\label{sec:supp_labels}
The 8,138 free text reports of the TCGA-Reports corpus~\cite{kefeli2024tcgareports} with cached slides were extracted by gpt-6-luna with a 17 field schema. Every value carries a verbatim quote that must be a substring of the report, checked by a validator, and records with imperfect quotes (1 to 2 percent per batch) were repaired or dropped. An independent extraction by Claude Sonnet 5.5 of 320 reports stratified over the 32 projects agrees with it on 97 to 100 percent of the reports where both give a value for a categorical field (kappa: grade 0.95, lymphovascular invasion 0.96, perineural invasion 0.98, necrosis 0.86, precursor 0.91, mitotic activity 0.82, multifocality 0.88, lymphocytic infiltrate 0.78), with 8 contradictions in 1,100 field pairs. Disagreements are missingness in both directions. Field coverage over the reports: grade 67 percent, lymphovascular invasion 63, precursor 22, necrosis 20, multifocality 22, perineural invasion 18, mitotic activity 12, lymphocytic infiltrate 3. A report that does not mention a finding gives no question, so every slide level number is agreement with the report where the report speaks.

\section{Extended setup and results}
\label{sec:supp_extended}
This section holds the parts of the protocol and of the results that the main text states in short form.

\paragraph{Field question sets.} From NuCLS~\cite{amgad2022nucls} we use 1,744 fields of 124 TCGA-BRCA patients with nucleus classes, and BCSS~\cite{amgad2019bcss} gives the tissue classes of their context. From these we build 19 operators in three families: cell questions about one nucleus (class, super class, size bin, the tissue it lies in, and whether it is of a class named at runtime), region questions about a box (count bin, presence, class fraction bin, which of two classes has more cells, tissue presence, fraction and dominant tissue, mitotic figures, stromal TIL score bin~\cite{salgado2015tils}), and compositional questions (which of two regions has more cells of a class, TIL count in a compartment, TIL location, density comparison), 242,604 questions split by patient (87 train, 19 validation, 18 test). PanNuke~\cite{gamper2020pannuke} adds 7,901 fields of 19 organs with 9 operators (781,295 questions, the official folds, and an organ split that holds out breast, colon, lung and prostate), trained jointly with NuCLS.

\paragraph{Slide question sets.} The pan cancer set covers 10,712 diagnostic slides of 8,861 patients over 32 TCGA projects~\cite{weinstein2013tcga}. Eight report fields shared across cancers (grade, lymphovascular invasion, perineural invasion, necrosis, precursor or in situ component, mitotic activity, multifocality, lymphocytic infiltrate) were extracted from 8,138 reports of the TCGA-Reports corpus~\cite{kefeli2024tcgareports} by gpt-6-luna~\cite{openai2026models} with a schema of verbatim quotes and audited against an independent extraction of 320 stratified reports (kappa 0.78 to 0.98 on the categorical fields, \cref{sec:supp_labels}). Cancer type (32 way), molecular subtype, tumour purity bins from the PanCanAtlas ABSOLUTE calls~\cite{carter2012absolute} and leukocyte fraction bins from the methylation estimate~\cite{thorsson2018immune} come from the clinical tables. Five composite operators are built by rule from report fields and held out of training for every \method model: high grade with LVI and necrosis with high mitotic activity from two of the pan cancer fields each, the precursor type from the type the report names when a precursor is present, and, on breast slides, the Nottingham grade implied by the tubule, pleomorphism and mitotic scores and DCIS with comedo necrosis from the synoptic report. A report belongs to a patient, and its fields are assigned to every diagnostic slide of that patient. Slides are split per project 70/15/15 by patient (6,772 / 1,402 / 1,393 slides) and four projects (KIRP, STAD, CESC, PAAD) are held out entirely (1,145 slides). The breast set adds the full synoptic report of 1,122 TCGA-BRCA slides (18 fields with the Nottingham components~\cite{elston1991nottingham}, DCIS, LCIS and calcifications, and ER, PR, HER2 and PAM50~\cite{parker2009pam50} from the molecular tables), split 780 / 172 / 170. Every report derived number is agreement with the report.

\paragraph{Splits.} The field split and the slide split are drawn separately, each by patient. The slide decoder and the per tile heads are trained on the slide training split only, and the field decoder on the field training split only (13 of its 87 patients have a slide in the slide test split, one of them among the 29 slides of the organiser study). Without the questions about patients of the field training or validation split (56 of the 995 composite questions, 8 of the 189 breast questions) the organiser is at 0.755 on the composites in place of 0.757, and each of the four models asking on the breast questions moves by 0.4 points.

\paragraph{Organiser study.} A TCGA barcode names the contributing site and thereby the cancer type, so the slide is named by a hash and the tools return no project, except in two sets of runs that keep the barcode and in which \texttt{slide\_info} returns the project: the conditions of the controlled study with tools or images other than free viewing by Claude Opus 5.5, and Claude Opus 5.5 asking the seven report fields of \cref{tab:ways_pan}. Asked both ways on the pan cancer split, Claude Opus 5.5 with the tools reaches 0.721 with the barcode and 0.712 with the hash on 1,484 synoptic questions, with the same answer on 97.4 percent, and 0.773 against 0.757 on the 995 composites. Through the API the tools conditions take 1.4k to 1.7k input tokens and 1.1 to 1.7 tool calls per question against 3.0k tokens when the model looks, and 6.7k against 50k through the Claude Code CLI, which reads the images as files. Looking, gpt-6-sol under calls necrosis (11 of 13 slides answered no when 10 are yes) and defaults to grade 2. Giving the model the images on top of the tools does not help (0.767 with both, 0.762 when left free to use any tool), and left free it called the tools on 166 of 189 questions and looked at tiles on 23, always together with a tool call. Claude Opus 5.5 and gpt-6-sol with the tools differ on one question of 189.

\paragraph{Slide level comparables.} TITAN encodes slides from CONCH v1.5 features at 512 px, PRISM2 from Virchow2 class tokens, \methS from Virchow2 tokens through a question decoder. \Cref{fig:pan_dots}a draws the pan cancer fields for every model: the probes of TITAN and PRISM2 and \methS land in the same place on the report fields and on the clinical table fields. CLAM-SB and a linear probe on the mean Virchow2 vector of the slide land in the same range as ABMIL (\cref{tab:pan}). Qwen3-VL-8B and Patho-R1-7B reading a thumbnail and six tiles read the 32 way cancer type at the majority class. The report PRISM2 writes states seven of the eight report fields for 26 to 69 percent of the slides (grade 49, lymphovascular invasion 61, necrosis 31), and where it states one it agrees with the reference at 0.60 to 0.80. It states the lymphocytic infiltrate for 4 of the 40 slides, too few to score. With an unstated field counted as wrong it is below the majority class on every report field (\cref{tab:pan}).

\paragraph{CAMELYON16.} The field decoder was trained on breast tumour fields and PanNuke. The slide score reads every tissue field on an 80 um grid (34k fields per slide), and its aggregation, the mean of the five highest probabilities, was chosen on a balanced subset of the training pool without any training. Half of the lesion fields are answered with zero tumour cells while every lesion free field is answered with zero. TITAN prompted with text separates all positives from negatives at 0.825 and Claude Opus 5.5 reading a thumbnail and six tiles at 0.680.

\subsection{Cost and confidence}
\label{sec:supp_cost}
\input{fig/fig6_cost_confidence}
\Cref{fig:cost}a plots the cost of a question against the number of questions asked of one state. \methS answers 32 questions in 2.9 ms after the 42 ms of encoding, 0.1 ms per question, 32 times faster than serial execution. On a slide the decoder answers 32 questions about the 4,096 sampled tiles in 2.6 ms, against 85 ms one by one, and one call of the organiser's slide tool with twelve operators takes 0.8 s, of which 0.7 s is the read of the cached tile vectors from disk (medians over 20 test slides). \methV with the shared prefix answers at 8.0 ms per question at 32 questions and 6.4 ms at 64, 8.5 and 10.9 times faster than recomputing the prefix. An 896 px state (four times the tokens) costs 5.4 times the encoding and changes accuracy by 0.001.

\Cref{fig:cost}b uses the probabilities for abstention. Abstaining on the 20 percent least confident questions~\cite{geifman2017selective} lifts \methS from 0.791 to 0.876 and the fusion from 0.805 to 0.891, and the fusion is the best calibrated reading (ECE 0.026). On PanNuke organs held out from training (\cref{fig:cost}c) accuracy drops from 0.842 to 0.783, the mean confidence of a field flags an unseen organ at AUROC 0.69, and abstaining on half of the questions recovers 0.928. At slide level the expected calibration error of \methS on the pan cancer test split is 0.052 over the trained operators and 0.050 over the eight report fields (0.03 to 0.14 per operator, three seeds), and abstaining on the least confident half of the report field questions lifts their accuracy from 0.705 to 0.830. The per tile heads with log mean exp pooling are at 0.043 (0.05 to 0.06 per finding, three seeds), and the confidence Claude Opus 5.5 states with its composed answers is at 0.039 on the 995 composite questions.

\subsection{Ablations in full}
\label{sec:supp_ablations}
\input{tables/T3_ablations}
\Cref{tab:ablations} changes one factor of \methS at a time. Four parts of the decoder matter. Without the box pooled query tokens accuracy drops by 0.021 and the comparison of two regions by 0.29, so the decoder needs a pooled summary of the box to compare two of them. Replacing the question conditioned attention by one fixed linear head per operator and argument pair costs 0.015 overall and 0.25 on nuclear size. Training on NuCLS alone instead of jointly with PanNuke costs 0.051 on the cell family and 0.25 on nuclear size. Removing the context canvas costs 0.015 and the region family 0.014. Nothing that enlarges the model moves accuracy by a point. A decoder of 768 dimensions and three plus three layers, 128 instead of 32 questions per step and the operator encoded by text each move the likelihood by 0.03 to 0.04 on one seed, about the seed spread of the reference (test NLL 0.529 $\pm$ 0.028 over three seeds), and accuracy by less than one point, the 896 px state gives NLL 0.532 against 0.529, and the last four Virchow2 blocks unfrozen give 0.529 against 0.549 for the frozen control run with the same schedule, with validation likelihood never improving. At slide level a decoder over every tile (NLL 0.801 and 0.809 on two seeds) and one over 1,024 sampled tiles (0.800) match the reference with 4,096 sampled tiles (0.815), and the decoder over every tile reads the cancer type one to two points worse. \methV reading only the pooled Virchow2 tokens of the state instead of its own image tokens reaches the same likelihood as the pixel path (0.563 against 0.563), so its accuracy comes from the pathology state and not from the base model's vision. At this scale of labels the encoder, the state resolution and the decoder are not the lever, the labels are, and the parts that matter are the ones that let a question address its box, its argument and its context.

\section{Full benchmarks}
\label{sec:supp_tables}
\Cref{tab:main_both,tab:main_sol} repeat \cref{tab:main} with gpt-6-sol looking at the same images, beside Claude Opus 5.5 and in its place. The remaining tables are the full benchmarks behind the sample of \cref{tab:main}. \Cref{tab:ways} lists every condition of the controlled study on the breast questions, in which a frontier model reads a description, looks or asks. Its grey image row is the control of the look condition: with grey images in place of the thumbnail and the six tiles gpt-6-sol answers at 0.460, so the pixels are worth 12.2 points to it (95 percent interval 4.0 to 20.1) and asking is worth 32.8 (20.7 to 44.8). \Cref{tab:ways_pan} repeats asking and reading on the whole pan cancer test split, 3,797 questions on 1,179 slides. gpt-6-sol answers at 0.722 asking \methS and at 0.637 reading the PRISM2 report, 8.6 points apart (6.6 to 10.5), and Claude Opus 5.5 asking answers at 0.731. Asking leads on six of the seven report fields, by 30.8 points on necrosis and 23.5 on precursor lesions, where the report stays below the majority class. On the composites the gap is 3.2 points (0.3 to 6.1), and reading is ahead on grade from its components (0.755 against 0.663).
\input{tables/T0_main_method_A}
\input{tables/T0_main_method_B}
\input{tables/T0_main}
\input{tables/T1_fov}
\input{tables/T2_slide_brca}
\input{tables/T5_slide_pan}
\input{tables/T6_slide_pan_heldout}
\input{tables/T4_router}
\input{tables/T7_ways}
\input{tables/T8_ways_pan}

\section{Additional figures}
\label{sec:supp_figs}
\Cref{fig:fov} draws the field benchmark of \cref{tab:fov} per operator. \Cref{fig:organiser_dots} draws the controlled study of \cref{sec:slide} per operator for six conditions. \Cref{fig:pan_dots} draws the pan cancer fields and the composites for every comparable of \cref{tab:pan}, with the four held out projects of \cref{tab:pan_heldout}. \Cref{fig:transfer_full,fig:camelyon} are the panels of \cref{fig:transfer} at full size, with a micro metastasis beside the macro metastasis.
\input{fig/figS_fov_radar}
\input{fig/fig5_transfer}
\input{fig/fig5b_camelyon_maps}
\input{fig/fig3_organiser}
\input{fig/fig4_pan_comparables}

\section{Figure assets}
\label{sec:supp_assets}
The H\&E images in \cref{fig:teaser,fig:method} are a TCGA-BRCA diagnostic slide and a NuCLS field with its context canvas. Icons are from Font Awesome Free (CC BY 4.0) and from Bioicons contributors (Servier, CC BY 3.0, Noelia Ferruz, CC BY 4.0, and CC0 contributors).

%% file: fig/fig6_cost_confidence.tex
\begin{figure*}[t]
\centering
\includegraphics[width=\textwidth]{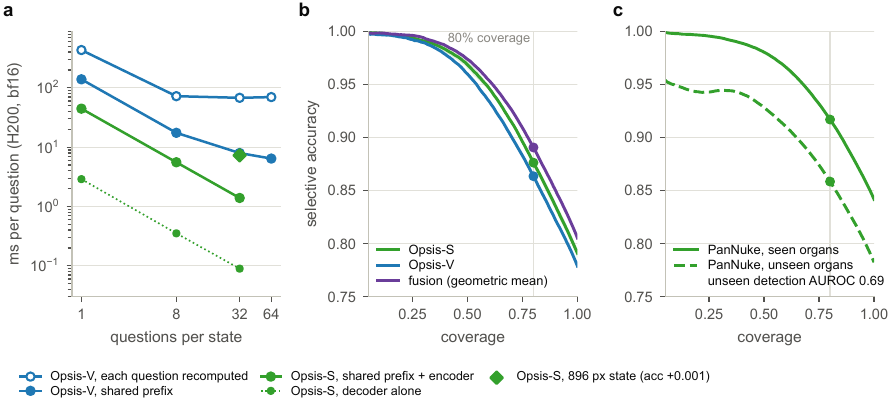}
\caption{\textbf{Cost and confidence.} (a) Milliseconds per question against questions per state on one H200 in bf16: \methV with every question recomputed or with the shared prefix, and \methS with the shared prefix including the frozen encoder once per state (42 ms at 448 px) or the decoder alone (0.1 ms per question at 32 questions). The 896 px state costs 5.4 $\times$ more encoding for the same accuracy. (b) Selective accuracy against coverage on the NuCLS test split (3 seeds, the band is one standard deviation): abstaining on the 20 percent least confident questions lifts \methS from 0.79 to 0.88 and the fusion of both instantiations to 0.89. (c) \methS on PanNuke organs seen in training against organs held out from training: accuracy drops from 0.84 to 0.78, and the mean confidence of a field flags an unseen organ with AUROC 0.69.}
\label{fig:cost}
\end{figure*}

%% file: tables/T3_ablations.tex
\begin{table*}[t]
\centering
\footnotesize\setlength{\tabcolsep}{3.5pt}
\caption{Ablations of Opsis-S on the NuCLS test split (each row changes one factor of the reference Opsis-S, the columns are the difference to the reference run of the same seed, mean over seeds) and the negative results. The context canvas, box pooled query tokens, question conditioning and joint PanNuke training matter. Decoder size, questions per step, query encoding, state resolution, a tuned encoder and a full bag decoder do not. The cell and composite columns are the operator families.}
\label{tab:ablations}
\begin{tabular}{lrrrrrr}
\toprule
ablation & seeds & $\Delta$ NLL $\downarrow$ & $\Delta$ acc $\uparrow$ & $\Delta$ cell $\uparrow$ & $\Delta$ composite $\uparrow$ & $\Delta$ PanNuke $\uparrow$ \\
\midrule
no context canvas & 3 & +0.026 & -0.015 & -0.022 & +0.013 & +0.000 \\
no box pooled query tokens & 3 & +0.041 & -0.021 & -0.024 & -0.071 & -0.062 \\
no density auxiliary loss & 3 & +0.010 & +0.004 & +0.010 & +0.027 & +0.002 \\
NuCLS only, no joint PanNuke & 3 & +0.018 & -0.008 & -0.051 & -0.004 &  \\
fixed multitask heads, no question conditioning & 3 & +0.035 & -0.015 & -0.061 & -0.024 & -0.031 \\
decoder d 192, 1 + 1 layers & 3 & +0.016 & -0.003 & -0.003 & -0.024 & -0.006 \\
8 questions per state per step & 3 & -0.002 & +0.003 & +0.001 & +0.016 & -0.001 \\
896 px state (4 x tokens, 5.4 x encoder cost) & 3 & +0.002 & +0.001 & +0.013 & +0.008 & +0.004 \\
decoder d 768, 3 + 3 layers & 1 & -0.031 & +0.009 & +0.030 & +0.003 & -0.002 \\
128 questions per state per step & 1 & -0.037 & +0.006 & +0.023 & -0.001 & -0.006 \\
operator encoded by text & 1 & -0.035 & +0.006 & +0.024 & +0.026 & +0.001 \\
compositional operators held out & 1 & -0.035 & +0.010 & +0.023 & -0.016 & -0.000 \\
Virchow2 last 4 blocks tuned (control: frozen, same schedule) & 1 & -0.020 &  &  &  &  \\
slide decoder over every tile instead of K = 4,096 sampled (slide level) & 2 & -0.010 &  &  &  &  \\
\bottomrule
\end{tabular}
\end{table*}

%% file: tables/T0_main_method_A.tex
\begin{table*}[t]
\centering
\footnotesize\setlength{\tabcolsep}{5pt}
\caption{Variant A of \cref{tab:main}: gpt-6-sol looking at the same images is added beside Claude Opus 5.5. gpt-6-sol sees what Opus 5.5 sees (the field and its context on the field rows, a 1024 px thumbnail and six 256 um tiles on the slide rows, one 128 um tile on the BCSS row), with reasoning at low effort. Rows, comparables and marking as in \cref{tab:main}.}
\label{tab:main_both}
\begin{tabular}{lrrrrrrrrr}
\toprule
& & \multicolumn{1}{c}{\makebox[0pt]{reference}} & \multicolumn{6}{c}{no labels for the task} & \multicolumn{1}{c}{ours} \\ \cmidrule(lr){3-3} \cmidrule(lr){4-9} \cmidrule(lr){10-10}
task & \rotatebox{90}{\textcolor[gray]{0.45}{maj.}} & \rotatebox{90}{\textcolor[gray]{0.45}{per task}} & \rotatebox{90}{TITAN zs} & \rotatebox{90}{PRISM2 zs} & \rotatebox{90}{Qwen3-VL} & \rotatebox{90}{Patho-R1} & \rotatebox{90}{gpt-6-sol} & \rotatebox{90}{Opus 5.5} & \rotatebox{90}{Opsis} \\
\midrule
\multicolumn{10}{l}{\textit{Field level, NuCLS test split (99 to 2,399 questions per row), accuracy. \method: fusion of \methS and \methV}} \\
cell super class & \textcolor[gray]{0.45}{0.47} & \textcolor[gray]{0.45}{0.804} &  &  & 0.471 & 0.484 & 0.659 & \underline{0.715} & \textbf{0.816} \\
cell class & \textcolor[gray]{0.45}{0.47} & \textcolor[gray]{0.45}{0.727} &  &  & 0.186 & 0.462 & 0.591 & \underline{0.668} & \textbf{0.748} \\
nuclear size bin & \textcolor[gray]{0.45}{0.35} &  &  &  & 0.309 & 0.176 & 0.369 & \underline{0.432} & \textbf{0.834} \\
compare regions & \textcolor[gray]{0.45}{0.52} &  &  &  & 0.491 & 0.484 & 0.748 & \underline{0.811} & \textbf{0.839} \\
dominant tissue & \textcolor[gray]{0.45}{0.51} &  &  &  & 0.442 & 0.561 & 0.598 & \underline{0.607} & \textbf{0.688} \\
sTIL score bin & \textcolor[gray]{0.45}{0.43} &  &  &  & 0.152 & \underline{0.434} & 0.263 & 0.424 & \textbf{0.663} \\
\addlinespace[2pt]\multicolumn{10}{l}{\textit{Slide level, TCGA-BRCA test split (170 slides), accuracy. \method: \methS}} \\
ER positive & \textcolor[gray]{0.45}{0.72} & \textcolor[gray]{0.45}{0.858} & 0.463 & \underline{0.778} & 0.710 & 0.648 & 0.667 & 0.636 & \textbf{0.813} \\
histologic grade & \textcolor[gray]{0.45}{0.48} & \textcolor[gray]{0.45}{0.674} & 0.483 & \textbf{0.697} & 0.414 & 0.331 & 0.572 & 0.586 & \underline{0.690} \\
LVI & \textcolor[gray]{0.45}{0.70} & \textcolor[gray]{0.45}{0.730} & 0.400 & \textbf{0.727} & 0.300 & \underline{0.700} & 0.436 & 0.673 & 0.694 \\
PAM50 subtype & \textcolor[gray]{0.45}{0.48} & \textcolor[gray]{0.45}{0.671} & 0.235 & \underline{0.524} & 0.229 & 0.482 & 0.412 & 0.459 & \textbf{0.655} \\
\addlinespace[2pt]\multicolumn{10}{l}{\textit{Slide level, pan cancer test split (1,393 slides of 28 projects), accuracy. \method: \methS}} \\
cancer type (32) & \textcolor[gray]{0.45}{0.12} & \textcolor[gray]{0.45}{0.933} & \underline{0.877} & 0.707 & 0.112 & 0.123 & 0.471 & 0.659 & \textbf{0.937} \\
necrosis & \textcolor[gray]{0.45}{0.66} & \textcolor[gray]{0.45}{0.829} & 0.694 & \underline{0.746} & 0.607 & 0.467 & 0.507 & 0.558 & \textbf{0.831} \\
LVI & \textcolor[gray]{0.45}{0.60} & \textcolor[gray]{0.45}{0.686} & 0.528 & \underline{0.649} & 0.407 & 0.594 & 0.504 & 0.615 & \textbf{0.679} \\
PNI & \textcolor[gray]{0.45}{0.53} & \textcolor[gray]{0.45}{0.780} & 0.543 & \textbf{0.790} & 0.554 & 0.473 & 0.586 & 0.710 & \underline{0.771} \\
mitotic activity & \textcolor[gray]{0.45}{0.61} & \textcolor[gray]{0.45}{0.770} & 0.616 & 0.502 & 0.530 & 0.724 & 0.641 & \textbf{0.779} & \underline{0.771} \\
tumour purity & \textcolor[gray]{0.45}{0.27} & \textcolor[gray]{0.45}{0.464} & 0.281 & 0.260 & 0.260 & 0.251 & 0.333 & \underline{0.339} & \textbf{0.468} \\
\addlinespace[2pt]\multicolumn{10}{l}{\textit{Composites held out of training (995 questions on 688 test slides), accuracy. \method: Opus 5.5 asking \methS}} \\
high grade with LVI & \textcolor[gray]{0.45}{0.76} & \textcolor[gray]{0.45}{0.781} & 0.451 & 0.503 & 0.284 & \underline{0.735} & 0.635 & 0.718 & \textbf{0.797} \\
precursor type & \textcolor[gray]{0.45}{0.50} & \textcolor[gray]{0.45}{0.784} & \underline{0.612} & 0.505 & 0.198 & 0.559 & \textbf{0.685} & 0.608 & \textbf{0.685} \\
necrosis with high mitotic & \textcolor[gray]{0.45}{0.71} & \textcolor[gray]{0.45}{0.879} & 0.559 & \textbf{0.842} & 0.404 & 0.815 & 0.794 & 0.815 & \underline{0.822} \\
grade from components & \textcolor[gray]{0.45}{0.51} & \textcolor[gray]{0.45}{0.694} & 0.531 & 0.653 & 0.347 & 0.429 & 0.612 & \textbf{0.694} & \underline{0.673} \\
DCIS with comedo necrosis & \textcolor[gray]{0.45}{0.74} & \textcolor[gray]{0.45}{0.713} & 0.620 & 0.340 & \textbf{0.780} & 0.340 & 0.560 & 0.320 & \underline{0.660} \\
all 995 composite questions & \textcolor[gray]{0.45}{0.67} & \textcolor[gray]{0.45}{0.784} & 0.519 & 0.560 & 0.314 & 0.657 & 0.663 & \underline{0.685} & \textbf{0.757} \\
\addlinespace[2pt]\multicolumn{10}{l}{\textit{Transfer without training on the dataset, AUROC. \method: \methS}} \\
CAMELYON16 metastasis, 129 slides &  &  & 0.825 & \textbf{0.968} & 0.484 & 0.511 & 0.580 & 0.680 & \underline{0.864} \\
BCSS necrosis, 8,830 tiles &  &  &  &  & 0.780 & \underline{0.958} & 0.908 & \textbf{0.970} & 0.933 \\
\bottomrule
\end{tabular}
\end{table*}

%% file: tables/T0_main_method_B.tex
\begin{table*}[t]
\centering
\footnotesize\setlength{\tabcolsep}{5.5pt}
\caption{Variant B of \cref{tab:main}: gpt-6-sol looking at the same images replaces Claude Opus 5.5 as the frontier vision language model. Input and marking as in \cref{tab:main_both}.}
\label{tab:main_sol}
\begin{tabular}{lrrrrrrrr}
\toprule
& & \multicolumn{1}{c}{\makebox[0pt]{reference}} & \multicolumn{5}{c}{no labels for the task} & \multicolumn{1}{c}{ours} \\ \cmidrule(lr){3-3} \cmidrule(lr){4-8} \cmidrule(lr){9-9}
task & \rotatebox{90}{\textcolor[gray]{0.45}{maj.}} & \rotatebox{90}{\textcolor[gray]{0.45}{per task}} & \rotatebox{90}{TITAN zs} & \rotatebox{90}{PRISM2 zs} & \rotatebox{90}{Qwen3-VL} & \rotatebox{90}{Patho-R1} & \rotatebox{90}{gpt-6-sol} & \rotatebox{90}{Opsis} \\
\midrule
\multicolumn{9}{l}{\textit{Field level, NuCLS test split (99 to 2,399 questions per row), accuracy. \method: fusion of \methS and \methV}} \\
cell super class & \textcolor[gray]{0.45}{0.47} & \textcolor[gray]{0.45}{0.804} &  &  & 0.471 & 0.484 & \underline{0.659} & \textbf{0.816} \\
cell class & \textcolor[gray]{0.45}{0.47} & \textcolor[gray]{0.45}{0.727} &  &  & 0.186 & 0.462 & \underline{0.591} & \textbf{0.748} \\
nuclear size bin & \textcolor[gray]{0.45}{0.35} &  &  &  & 0.309 & 0.176 & \underline{0.369} & \textbf{0.834} \\
compare regions & \textcolor[gray]{0.45}{0.52} &  &  &  & 0.491 & 0.484 & \underline{0.748} & \textbf{0.839} \\
dominant tissue & \textcolor[gray]{0.45}{0.51} &  &  &  & 0.442 & 0.561 & \underline{0.598} & \textbf{0.688} \\
sTIL score bin & \textcolor[gray]{0.45}{0.43} &  &  &  & 0.152 & \underline{0.434} & 0.263 & \textbf{0.663} \\
\addlinespace[2pt]\multicolumn{9}{l}{\textit{Slide level, TCGA-BRCA test split (170 slides), accuracy. \method: \methS}} \\
ER positive & \textcolor[gray]{0.45}{0.72} & \textcolor[gray]{0.45}{0.858} & 0.463 & \underline{0.778} & 0.710 & 0.648 & 0.667 & \textbf{0.813} \\
histologic grade & \textcolor[gray]{0.45}{0.48} & \textcolor[gray]{0.45}{0.674} & 0.483 & \textbf{0.697} & 0.414 & 0.331 & 0.572 & \underline{0.690} \\
LVI & \textcolor[gray]{0.45}{0.70} & \textcolor[gray]{0.45}{0.730} & 0.400 & \textbf{0.727} & 0.300 & \underline{0.700} & 0.436 & 0.694 \\
PAM50 subtype & \textcolor[gray]{0.45}{0.48} & \textcolor[gray]{0.45}{0.671} & 0.235 & \underline{0.524} & 0.229 & 0.482 & 0.412 & \textbf{0.655} \\
\addlinespace[2pt]\multicolumn{9}{l}{\textit{Slide level, pan cancer test split (1,393 slides of 28 projects), accuracy. \method: \methS}} \\
cancer type (32) & \textcolor[gray]{0.45}{0.12} & \textcolor[gray]{0.45}{0.933} & \underline{0.877} & 0.707 & 0.112 & 0.123 & 0.471 & \textbf{0.937} \\
necrosis & \textcolor[gray]{0.45}{0.66} & \textcolor[gray]{0.45}{0.829} & 0.694 & \underline{0.746} & 0.607 & 0.467 & 0.507 & \textbf{0.831} \\
LVI & \textcolor[gray]{0.45}{0.60} & \textcolor[gray]{0.45}{0.686} & 0.528 & \underline{0.649} & 0.407 & 0.594 & 0.504 & \textbf{0.679} \\
PNI & \textcolor[gray]{0.45}{0.53} & \textcolor[gray]{0.45}{0.780} & 0.543 & \textbf{0.790} & 0.554 & 0.473 & 0.586 & \underline{0.771} \\
mitotic activity & \textcolor[gray]{0.45}{0.61} & \textcolor[gray]{0.45}{0.770} & 0.616 & 0.502 & 0.530 & \underline{0.724} & 0.641 & \textbf{0.771} \\
tumour purity & \textcolor[gray]{0.45}{0.27} & \textcolor[gray]{0.45}{0.464} & 0.281 & 0.260 & 0.260 & 0.251 & \underline{0.333} & \textbf{0.468} \\
\addlinespace[2pt]\multicolumn{9}{l}{\textit{Composites held out of training (995 questions on 688 test slides), accuracy. \method: Opus 5.5 asking \methS}} \\
high grade with LVI & \textcolor[gray]{0.45}{0.76} & \textcolor[gray]{0.45}{0.781} & 0.451 & 0.503 & 0.284 & \underline{0.735} & 0.635 & \textbf{0.797} \\
precursor type & \textcolor[gray]{0.45}{0.50} & \textcolor[gray]{0.45}{0.784} & \underline{0.612} & 0.505 & 0.198 & 0.559 & \textbf{0.685} & \textbf{0.685} \\
necrosis with high mitotic & \textcolor[gray]{0.45}{0.71} & \textcolor[gray]{0.45}{0.879} & 0.559 & \textbf{0.842} & 0.404 & 0.815 & 0.794 & \underline{0.822} \\
grade from components & \textcolor[gray]{0.45}{0.51} & \textcolor[gray]{0.45}{0.694} & 0.531 & \underline{0.653} & 0.347 & 0.429 & 0.612 & \textbf{0.673} \\
DCIS with comedo necrosis & \textcolor[gray]{0.45}{0.74} & \textcolor[gray]{0.45}{0.713} & 0.620 & 0.340 & \textbf{0.780} & 0.340 & 0.560 & \underline{0.660} \\
all 995 composite questions & \textcolor[gray]{0.45}{0.67} & \textcolor[gray]{0.45}{0.784} & 0.519 & 0.560 & 0.314 & 0.657 & \underline{0.663} & \textbf{0.757} \\
\addlinespace[2pt]\multicolumn{9}{l}{\textit{Transfer without training on the dataset, AUROC. \method: \methS}} \\
CAMELYON16 metastasis, 129 slides &  &  & 0.825 & \textbf{0.968} & 0.484 & 0.511 & 0.580 & \underline{0.864} \\
BCSS necrosis, 8,830 tiles &  &  &  &  & 0.780 & \textbf{0.958} & 0.908 & \underline{0.933} \\
\bottomrule
\end{tabular}
\end{table*}

%% file: tables/T0_main.tex
\begin{table*}[t]
\centering
\scriptsize\setlength{\tabcolsep}{2pt}
\caption{The sample of \cref{tab:main} with one column per \method component. The full benchmarks are in \cref{tab:fov,tab:slide,tab:pan,tab:pan_heldout}. As in \cref{tab:main}, bold is the best and underline the second best outside the grey columns, the majority class and the references trained per task with labels. CellViT++ is the Virchow variant trained on NuCLS, ABMIL is the gated attention MIL of Ilse et al. and CLAM-SB the same network with the instance clustering loss, both trained per field on our split with labels, attn. MIL is our tile head with gated attention pooling on the same report labels, and Virchow2 pr. is a linear probe on the frozen encoder under \method, trained per task with labels: on the mean of the Virchow2 tile vectors of the slide, and on the field rows on the question's box pooled from the same tokens \methS reads. The label free comparables are TITAN and PRISM2 prompted with text (zero shot, zs), and Qwen3-VL-8B, Patho-R1-7B and Claude Opus 5.5 (thinking on at low effort) answering the typed questions from the field and its context on the field rows and from a thumbnail and six tiles on the slide rows (the pretraining data of Patho-R1 include PathGen-1.6M, captions generated on TCGA tiles, so it is not zero shot on the tissue). TITAN and PRISM2 align slide embeddings with pathology reports in pretraining, so their linear probes trained on our labels are kept to \cref{tab:slide,tab:pan}. Opsis-S is the question decoder over the state (the field decoder, or the slide decoder with K sampled tiles), tile heads its full bag read on the fields that have one, Opsis-V the Qwen3-VL-8B LoRA, fusion the geometric mean of the two probability vectors (the default answer when both run, \cref{tab:router}), organiser Claude Opus 5.5 calling Opsis-S as tools. The composite rows score every model on the 995 composite questions of the pan cancer test split: CLAM-SB is trained on the composite labels, TITAN and PRISM2 are asked the composite in text, the organiser composes the answer from the primitive tools, and the Opsis-S column is the slide decoder with a fixed rule over its primitive probabilities (\cref{sec:supp_organiser}), blank on precursor type, which has no rule and counts as wrong in the last row, and a zero shot cell is blank where the model has no prompt for the composite. On the slide rows the vision language models see a 1024 px thumbnail and six 256 um tiles. The CAMELYON16 row is the official 129 slide test set, where the best challenge entry trained on the cohort reaches 0.994.}
\label{tab:main_components}
\begin{tabular}{lrrrrrrrrrrrrrrrr}
\toprule
& & \multicolumn{5}{c}{reference: trained per task} & \multicolumn{5}{c}{no labels for the task} & \multicolumn{5}{c}{\method} \\ \cmidrule(lr){3-7} \cmidrule(lr){8-12} \cmidrule(lr){13-17}
task & \rotatebox{90}{\textcolor[gray]{0.45}{maj.}} & \rotatebox{90}{\textcolor[gray]{0.45}{CellViT++}} & \rotatebox{90}{\textcolor[gray]{0.45}{ABMIL}} & \rotatebox{90}{\textcolor[gray]{0.45}{CLAM-SB}} & \rotatebox{90}{\textcolor[gray]{0.45}{Virchow2 pr.}} & \rotatebox{90}{\textcolor[gray]{0.45}{attn. MIL}} & \rotatebox{90}{TITAN zs} & \rotatebox{90}{PRISM2 zs} & \rotatebox{90}{Qwen3-VL} & \rotatebox{90}{Patho-R1} & \rotatebox{90}{Opus 5.5} & \rotatebox{90}{Opsis-S} & \rotatebox{90}{tile heads} & \rotatebox{90}{Opsis-V} & \rotatebox{90}{fusion} & \rotatebox{90}{organiser} \\
\midrule
\multicolumn{17}{l}{\textit{Field level, NuCLS test split (99 to 2,399 questions per row), accuracy}} \\
cell super class & \textcolor[gray]{0.45}{0.47} & \textcolor[gray]{0.45}{0.804} &  &  & \textcolor[gray]{0.45}{0.797} &  &  &  & 0.471 & 0.484 & 0.715 & \underline{0.800} &  & 0.766 & \textbf{0.816} &  \\
cell class & \textcolor[gray]{0.45}{0.47} & \textcolor[gray]{0.45}{0.727} &  &  & \textcolor[gray]{0.45}{0.720} &  &  &  & 0.186 & 0.462 & 0.668 & \underline{0.739} &  & 0.687 & \textbf{0.748} &  \\
nuclear size bin & \textcolor[gray]{0.45}{0.35} &  &  &  & \textcolor[gray]{0.45}{0.510} &  &  &  & 0.309 & 0.176 & 0.432 & 0.758 &  & \underline{0.802} & \textbf{0.834} &  \\
compare regions & \textcolor[gray]{0.45}{0.52} &  &  &  & \textcolor[gray]{0.45}{0.736} &  &  &  & 0.491 & 0.484 & \underline{0.811} & 0.784 &  & 0.801 & \textbf{0.839} &  \\
dominant tissue & \textcolor[gray]{0.45}{0.51} &  &  &  & \textcolor[gray]{0.45}{0.615} &  &  &  & 0.442 & 0.561 & 0.607 & \underline{0.642} &  & \textbf{0.688} & \textbf{0.688} &  \\
sTIL score bin & \textcolor[gray]{0.45}{0.43} &  &  &  & \textcolor[gray]{0.45}{0.717} &  &  &  & 0.152 & 0.434 & 0.424 & \underline{0.650} &  & \underline{0.650} & \textbf{0.663} &  \\
\addlinespace[2pt]\multicolumn{17}{l}{\textit{Slide level, TCGA-BRCA test split (170 slides), accuracy}} \\
ER positive & \textcolor[gray]{0.45}{0.72} &  & \textcolor[gray]{0.45}{0.858} & \textcolor[gray]{0.45}{0.850} & \textcolor[gray]{0.45}{0.833} &  & 0.463 & \underline{0.778} & 0.710 & 0.648 & 0.636 & \textbf{0.813} &  &  &  &  \\
histologic grade & \textcolor[gray]{0.45}{0.48} &  & \textcolor[gray]{0.45}{0.674} & \textcolor[gray]{0.45}{0.671} & \textcolor[gray]{0.45}{0.690} &  & 0.483 & \textbf{0.697} & 0.414 & 0.331 & 0.586 & \underline{0.690} &  &  &  &  \\
LVI & \textcolor[gray]{0.45}{0.70} &  & \textcolor[gray]{0.45}{0.730} & \textcolor[gray]{0.45}{0.730} & \textcolor[gray]{0.45}{0.691} &  & 0.400 & \textbf{0.727} & 0.300 & \underline{0.700} & 0.673 & 0.694 &  &  &  &  \\
PAM50 subtype & \textcolor[gray]{0.45}{0.48} &  & \textcolor[gray]{0.45}{0.671} & \textcolor[gray]{0.45}{0.698} & \textcolor[gray]{0.45}{0.653} &  & 0.235 & \underline{0.524} & 0.229 & 0.482 & 0.459 & \textbf{0.655} &  &  &  &  \\
\addlinespace[2pt]\multicolumn{17}{l}{\textit{Slide level, pan cancer test split (1,393 slides of 28 projects), accuracy}} \\
cancer type (32) & \textcolor[gray]{0.45}{0.12} &  & \textcolor[gray]{0.45}{0.933} & \textcolor[gray]{0.45}{0.936} & \textcolor[gray]{0.45}{0.938} &  & \underline{0.877} & 0.707 & 0.112 & 0.123 & 0.659 & \textbf{0.937} &  &  &  &  \\
necrosis & \textcolor[gray]{0.45}{0.66} &  & \textcolor[gray]{0.45}{0.829} & \textcolor[gray]{0.45}{0.812} & \textcolor[gray]{0.45}{0.818} &  & 0.694 & 0.746 & 0.607 & 0.467 & 0.558 & \underline{0.821} & \textbf{0.831} &  &  &  \\
LVI & \textcolor[gray]{0.45}{0.60} &  & \textcolor[gray]{0.45}{0.686} & \textcolor[gray]{0.45}{0.688} & \textcolor[gray]{0.45}{0.676} &  & 0.528 & \underline{0.649} & 0.407 & 0.594 & 0.615 & 0.637 & \textbf{0.679} &  &  &  \\
PNI & \textcolor[gray]{0.45}{0.53} &  & \textcolor[gray]{0.45}{0.780} & \textcolor[gray]{0.45}{0.758} & \textcolor[gray]{0.45}{0.774} &  & 0.543 & \textbf{0.790} & 0.554 & 0.473 & 0.710 & 0.746 & \underline{0.771} &  &  &  \\
mitotic activity & \textcolor[gray]{0.45}{0.61} &  & \textcolor[gray]{0.45}{0.770} & \textcolor[gray]{0.45}{0.774} & \textcolor[gray]{0.45}{0.774} &  & 0.616 & 0.502 & 0.530 & 0.724 & \textbf{0.779} & \underline{0.771} &  &  &  &  \\
tumour purity & \textcolor[gray]{0.45}{0.27} &  & \textcolor[gray]{0.45}{0.464} & \textcolor[gray]{0.45}{0.472} & \textcolor[gray]{0.45}{0.445} &  & 0.281 & 0.260 & 0.260 & 0.251 & \underline{0.339} & \textbf{0.468} &  &  &  &  \\
\addlinespace[2pt]\multicolumn{17}{l}{\textit{Composites held out of training (995 questions on 688 test slides), accuracy}} \\
high grade with LVI & \textcolor[gray]{0.45}{0.76} &  & \textcolor[gray]{0.45}{0.781} & \textcolor[gray]{0.45}{0.781} & \textcolor[gray]{0.45}{0.770} &  & 0.451 & 0.503 & 0.284 & 0.735 & 0.718 & \underline{0.782} &  &  &  & \textbf{0.797} \\
precursor type & \textcolor[gray]{0.45}{0.50} &  & \textcolor[gray]{0.45}{0.784} & \textcolor[gray]{0.45}{0.788} & \textcolor[gray]{0.45}{0.784} &  & \underline{0.612} & 0.505 & 0.198 & 0.559 & 0.608 &  &  &  &  & \textbf{0.685} \\
necrosis with high mitotic & \textcolor[gray]{0.45}{0.71} &  & \textcolor[gray]{0.45}{0.879} & \textcolor[gray]{0.45}{0.890} & \textcolor[gray]{0.45}{0.897} &  & 0.559 & \underline{0.842} & 0.404 & 0.815 & 0.815 & \textbf{0.890} &  &  &  & 0.822 \\
grade from components & \textcolor[gray]{0.45}{0.51} &  & \textcolor[gray]{0.45}{0.694} & \textcolor[gray]{0.45}{0.724} & \textcolor[gray]{0.45}{0.653} &  & 0.531 & 0.653 & 0.347 & 0.429 & \textbf{0.694} & \underline{0.680} &  &  &  & 0.673 \\
DCIS with comedo necrosis & \textcolor[gray]{0.45}{0.74} &  & \textcolor[gray]{0.45}{0.713} & \textcolor[gray]{0.45}{0.740} & \textcolor[gray]{0.45}{0.720} &  & 0.620 & 0.340 & \textbf{0.780} & 0.340 & 0.320 & 0.640 &  &  &  & \underline{0.660} \\
all 995 composite questions & \textcolor[gray]{0.45}{0.67} &  & \textcolor[gray]{0.45}{0.784} & \textcolor[gray]{0.45}{0.791} & \textcolor[gray]{0.45}{0.778} &  & 0.519 & 0.560 & 0.314 & 0.657 & \underline{0.685} & 0.606 &  &  &  & \textbf{0.757} \\
\addlinespace[2pt]\multicolumn{17}{l}{\textit{Transfer without training on the dataset, AUROC}} \\
CAMELYON16 metastasis, 129 slides &  &  &  &  &  &  & 0.825 & \textbf{0.968} & 0.484 & 0.511 & 0.680 &  & \underline{0.864} &  &  &  \\
BCSS necrosis, 8,830 tiles &  &  &  &  &  & \textcolor[gray]{0.45}{0.942} &  &  & 0.780 & \underline{0.958} & \textbf{0.970} &  & 0.933 &  &  &  \\
\bottomrule
\end{tabular}
\end{table*}

%% file: tables/T1_fov.tex
\begin{table*}[t]
\centering
\footnotesize\setlength{\tabcolsep}{3.5pt}
\caption{Field bench on the NuCLS test split (accuracy, mean $\pm$ sd over 3 seeds for Opsis-S and Opsis-V, mean over 3 seed pairs for their fusion). Virchow2 pr. is a linear probe on the frozen Virchow2 tokens Opsis-S reads (the question's box pooled as the decoder pools it, the second box, and the mean field token), one classifier per operator and argument group trained with labels, blank where the operator takes a runtime candidate or has more argument groups than the probe caps. The zero shot (zs) vision language models Qwen3-VL-8B and Patho-R1-7B answer the same typed questions from the same images as Opsis-V. Patho-R1 is not zero shot on TCGA tissue in the strict sense, its continued pretraining includes PathGen-1.6M, captions generated on TCGA tiles.}
\label{tab:fov}
\begin{tabular}{lrrrrrrrr}
\toprule
operator & n & Virchow2 pr. & majority & Opsis-S & Opsis-V & fusion & Qwen3-VL (zs) & Patho-R1 (zs) \\
\midrule
nuclear size bin & 857 & 0.510 & 0.35 & 0.758 $\pm$ 0.044 & 0.802 $\pm$ 0.018 & \textbf{0.834} & 0.309 & 0.176 \\
cell fraction bin & 2824 & 0.730 & 0.55 & 0.727 $\pm$ 0.015 & 0.712 $\pm$ 0.004 & \textbf{0.736} & 0.505 & 0.552 \\
cell is (runtime candidate) & 2399 &  & 0.51 & 0.875 $\pm$ 0.013 & 0.857 $\pm$ 0.006 & \textbf{0.889} & 0.506 & 0.581 \\
cell class & 2399 & 0.720 & 0.47 & 0.739 $\pm$ 0.015 & 0.687 $\pm$ 0.008 & \textbf{0.748} & 0.186 & 0.462 \\
cell super class & 2399 & 0.797 & 0.47 & 0.800 $\pm$ 0.008 & 0.766 $\pm$ 0.005 & \textbf{0.816} & 0.471 & 0.484 \\
cell's tissue & 1564 & 0.371 & 0.40 & 0.580 $\pm$ 0.011 & \textbf{0.612 $\pm$ 0.010} & \textbf{0.612} & 0.434 & 0.444 \\
compare regions (box, box) & 159 & 0.736 & 0.52 & 0.784 $\pm$ 0.020 & 0.801 $\pm$ 0.031 & \textbf{0.839} & 0.491 & 0.484 \\
count bin & 3408 & 0.721 & 0.58 & 0.735 $\pm$ 0.014 & 0.725 $\pm$ 0.011 & \textbf{0.750} & 0.302 & 0.576 \\
count in compartment & 181 & \textbf{0.630} & 0.48 & 0.619 $\pm$ 0.031 & 0.584 $\pm$ 0.008 & 0.608 & 0.470 & 0.475 \\
density compare & 41 & 0.463 & 0.54 & \textbf{0.512 $\pm$ 0.065} & 0.455 $\pm$ 0.075 & 0.480 & 0.488 & 0.488 \\
dominant tissue & 631 & 0.615 & 0.51 & 0.642 $\pm$ 0.021 & \textbf{0.688 $\pm$ 0.016} & \textbf{0.688} & 0.442 & 0.561 \\
mitotic count bin & 213 & 0.878 & 0.89 & 0.884 $\pm$ 0.003 & 0.884 $\pm$ 0.007 & \textbf{0.887} & 0.080 & 0.845 \\
mitotic figure present & 213 & 0.878 & 0.89 & \textbf{0.886 $\pm$ 0.007} & \textbf{0.886 $\pm$ 0.005} & \textbf{0.886} & 0.136 & 0.512 \\
more cells & 1868 & 0.872 & 0.59 & 0.902 $\pm$ 0.012 & 0.875 $\pm$ 0.008 & \textbf{0.908} & 0.585 & 0.561 \\
present & 3408 & 0.869 & 0.58 & 0.885 $\pm$ 0.008 & 0.864 $\pm$ 0.004 & \textbf{0.892} & 0.461 & 0.564 \\
sTIL score bin & 99 & \textbf{0.717} & 0.43 & 0.650 $\pm$ 0.015 & 0.650 $\pm$ 0.025 & 0.663 & 0.152 & 0.434 \\
TILs location & 12 & 0.667 & 0.67 & 0.722 $\pm$ 0.048 & \textbf{0.833 $\pm$ 0.000} & 0.750 & 0.667 & 0.667 \\
tissue fraction bin & 3030 & 0.651 & 0.57 & 0.727 $\pm$ 0.008 & 0.727 $\pm$ 0.004 & \textbf{0.738} & 0.254 & 0.572 \\
tissue present & 3030 & 0.912 & 0.57 & 0.929 $\pm$ 0.001 & 0.924 $\pm$ 0.001 & \textbf{0.934} & 0.646 & 0.669 \\
\bottomrule
\end{tabular}
\end{table*}

%% file: tables/T2_slide_brca.tex
\begin{table*}[t]
\centering
\footnotesize\setlength{\tabcolsep}{3.5pt}
\caption{Slide level bench on the TCGA-BRCA test split (170 slides, accuracy). Report fields are agreement with the report. Opsis-S is the slide decoder trained on the breast question set, 3 seeds. ABMIL (the gated attention MIL of Ilse et al., 3 seeds), CLAM-SB (seven fields), a linear probe on the mean Virchow2 tile vector of the slide (Virchow2 pr., the frozen encoder under Opsis-S) and the TITAN and PRISM2 linear probes (pr.) are trained on our split with labels. TITAN and PRISM2 align slide embeddings with report text in pretraining, so their probes are not label free in the sense of the zero shot columns. Zero shot (zs) prompts the slide foundation model with text, and the mismatch control gives PRISM2 zero shot another slide.}
\label{tab:slide}
\begin{tabular}{lrrrrrrrrrr}
\toprule
field & maj. & Opsis-S & ABMIL & CLAM-SB & Virchow2 pr. & TITAN pr. & PRISM2 pr. & TITAN zs & PRISM2 zs & mismatch \\
\midrule
ER positive & 0.72 & 0.813 & 0.858 & 0.850 & 0.833 & \textbf{0.864} & \textbf{0.864} & 0.463 & 0.778 & 0.605 \\
PR positive & 0.65 & 0.758 & \textbf{0.810} & 0.803 & 0.770 & 0.776 & 0.801 & 0.472 & 0.634 & 0.453 \\
HER2 positive & 0.87 & 0.876 & 0.876 & 0.878 & 0.876 & \textbf{0.883} & 0.832 & 0.460 & 0.869 & 0.869 \\
PAM50 subtype & 0.48 & 0.655 & 0.671 & \textbf{0.698} & 0.653 & 0.676 & 0.688 & 0.235 & 0.524 & 0.400 \\
histologic grade & 0.48 & 0.690 & 0.674 & 0.671 & 0.690 & \textbf{0.710} & 0.676 & 0.483 & 0.697 & 0.331 \\
histologic type & 0.68 & 0.803 & 0.824 & 0.826 & 0.758 & 0.826 & \textbf{0.832} & 0.727 & 0.795 & 0.528 \\
in situ component & 0.88 & 0.793 & 0.796 &  & 0.786 & 0.816 & \textbf{0.857} & 0.765 & 0.673 & 0.561 \\
LVI & 0.70 & 0.694 & 0.730 & 0.730 & 0.691 & \textbf{0.764} & 0.727 & 0.400 & 0.727 & 0.600 \\
necrosis & 0.82 & 0.702 & 0.786 &  & 0.750 & \textbf{0.821} & 0.750 & 0.750 & 0.714 & 0.500 \\
multifocal & 0.53 & 0.627 & 0.719 &  & 0.706 & \textbf{0.745} & 0.647 & 0.451 & 0.431 & 0.451 \\
LCIS & 0.70 & 0.773 & 0.752 &  & 0.766 & \textbf{0.809} & 0.723 & 0.723 & 0.723 & 0.340 \\
calcifications & 0.84 & 0.836 & 0.801 &  & \textbf{0.851} & \textbf{0.851} & 0.806 & 0.821 & 0.478 & 0.328 \\
DCIS comedo necrosis & 0.86 & \textbf{0.863} & 0.850 &  & \textbf{0.863} & \textbf{0.863} & \textbf{0.863} & 0.627 & 0.196 & 0.137 \\
nuclear grade & 0.51 & 0.694 & 0.694 &  & \textbf{0.750} & \textbf{0.750} & 0.583 & 0.583 & \textbf{0.750} & 0.250 \\
tubule score & 0.85 & 0.859 & 0.824 &  & 0.843 & \textbf{0.873} & \textbf{0.873} & 0.196 & 0.206 & 0.127 \\
pleomorphism score & 0.51 & 0.723 & 0.706 &  & 0.713 & 0.743 & \textbf{0.752} & 0.713 & 0.594 & 0.356 \\
mitotic score & 0.43 & 0.605 & 0.591 &  & 0.577 & 0.619 & \textbf{0.629} & 0.423 & 0.536 & 0.258 \\
DCIS grade & 0.52 & 0.689 & 0.622 &  & 0.617 & \textbf{0.817} & 0.683 & 0.533 & 0.733 & 0.350 \\
\bottomrule
\end{tabular}
\end{table*}

%% file: tables/T5_slide_pan.tex
\begin{table*}[t]
\centering
\footnotesize\setlength{\tabcolsep}{1.5pt}
\caption{Pan cancer slide bench on the pan cancer test split (1,393 slides of 28 TCGA projects, accuracy, the cancer type question has 32 candidates). Opsis-S is the pan cancer slide decoder and tile heads its full bag tile head pooled with log mean exp on the fields that have one, both the mean over 3 seeds. ABMIL is the gated attention MIL of Ilse et al. and CLAM-SB the same network with the instance clustering loss, both one classifier per field on 4,096 Virchow2 tiles per slide (ABMIL three seeds, CLAM-SB seed 0 and the test split only), Virchow2 pr. is a linear probe on the mean Virchow2 tile vector of the slide, the frozen encoder under Opsis-S, and the TITAN and PRISM2 linear probes (pr.) one classifier per field on their slide embeddings. All of them train on our split with labels, including the five composite fields below the rule, which Opsis-S was never trained on (the organiser composes them from the primitives, \cref{fig:pan_dots}). TITAN and PRISM2 align slide embeddings with report text in pretraining. Zero shot (zs) prompts the slide foundation model with text. The PRISM2 report column scores the free text report PRISM2 writes for the slide, extracted with the TCGA schema, with an unstated field counted as wrong. Majority is the majority class.}
\label{tab:pan}
\begin{tabular}{lrrrrrrrrrrr}
\toprule
field & maj. & Opsis-S & tile heads & ABMIL & CLAM-SB & Virchow2 pr. & TITAN pr. & PRISM2 pr. & TITAN zs & PRISM2 zs & PRISM2 report \\
\midrule
cancer type (32) & 0.12 & 0.937 &  & 0.933 & 0.936 & 0.938 & \textbf{0.949} & 0.942 & 0.877 & 0.707 &  \\
grade & 0.45 & 0.674 &  & 0.674 &  & 0.661 & 0.683 & \textbf{0.704} & 0.366 & 0.525 & 0.317 \\
LVI & 0.60 & 0.637 & 0.679 & 0.686 & \textbf{0.688} & 0.676 & 0.653 & 0.679 & 0.528 & 0.649 & 0.381 \\
PNI & 0.53 & 0.746 & 0.771 & 0.780 & 0.758 & 0.774 & 0.796 & \textbf{0.801} & 0.543 & 0.790 & 0.478 \\
necrosis & 0.66 & 0.821 & 0.831 & 0.829 & 0.812 & 0.818 & 0.847 & \textbf{0.849} & 0.694 & 0.746 & 0.185 \\
mitotic activity & 0.61 & 0.771 &  & 0.770 & 0.774 & 0.774 & \textbf{0.861} & 0.783 & 0.616 & 0.502 & 0.180 \\
multifocal & 0.56 & 0.626 &  & 0.671 &  & \textbf{0.676} & 0.654 & 0.660 & 0.467 & 0.569 & 0.229 \\
infiltrate & 0.45 & \textbf{0.642} &  & 0.575 &  & 0.575 & 0.500 & 0.375 & 0.400 & 0.450 & 0.050 \\
precursor present & 0.81 & 0.830 & 0.838 & \textbf{0.846} &  & 0.834 & 0.822 & 0.841 & 0.279 & 0.227 & 0.444 \\
molecular subtype & 0.31 & 0.584 &  &  &  & 0.581 & \textbf{0.631} & 0.619 &  &  &  \\
tumour purity & 0.27 & 0.468 &  & 0.464 & \textbf{0.472} & 0.445 & 0.441 & 0.436 & 0.281 & 0.260 &  \\
leukocyte fraction & 0.28 & 0.497 &  & 0.511 &  & 0.519 & \textbf{0.525} & 0.510 & 0.241 & 0.267 &  \\
mean of the 12 items & 0.471 & 0.686 &  &  &  & 0.689 & \textbf{0.697} & 0.683 &  &  &  \\
\midrule
DCIS with comedo necrosis & 0.74 & 0.660 &  & 0.713 & 0.740 & 0.720 & 0.720 & \textbf{0.780} & 0.620 & 0.340 &  \\
grade from components & 0.51 & 0.544 &  & 0.694 & 0.724 & 0.653 & \textbf{0.755} & 0.745 & 0.531 & 0.653 &  \\
high grade with LVI & 0.76 & 0.563 &  & 0.781 & 0.781 & 0.770 & 0.804 & \textbf{0.806} & 0.451 & 0.503 & 0.290 \\
necrosis with high mitotic & 0.71 & 0.758 &  & 0.879 & 0.890 & 0.897 & \textbf{0.952} & 0.925 & 0.559 & 0.842 & 0.068 \\
precursor type & 0.50 & 0.098 &  & 0.784 & 0.788 & 0.784 & 0.776 & \textbf{0.793} & 0.612 & 0.505 &  \\
\bottomrule
\end{tabular}
\end{table*}

%% file: tables/T6_slide_pan_heldout.tex
\begin{table*}[t]
\centering
\footnotesize\setlength{\tabcolsep}{3.5pt}
\caption{Pan cancer slide bench on the four held out projects (KIRP, STAD, CESC and PAAD, never seen in training by any method trained on our split). Cancer type and molecular subtype are unseen classes by construction. Columns as in \cref{tab:pan}. CLAM-SB is scored on the test split only and has no column here.}
\label{tab:pan_heldout}
\begin{tabular}{lrrrrrrrrrr}
\toprule
field & maj. & Opsis-S & tile heads & ABMIL & Virchow2 pr. & TITAN pr. & PRISM2 pr. & TITAN zs & PRISM2 zs & PRISM2 report \\
\midrule
cancer type (32) & 0.33 & 0.000 &  & 0.000 & 0.000 & 0.000 & 0.000 & \textbf{0.778} & 0.433 &  \\
grade & 0.46 & 0.567 &  & \textbf{0.604} & 0.568 & 0.540 & 0.601 & 0.300 & 0.518 & 0.360 \\
LVI & 0.51 & 0.627 & 0.664 & 0.687 & 0.675 & 0.677 & 0.691 & 0.551 & \textbf{0.701} & 0.401 \\
PNI & 0.68 & 0.630 & 0.577 & 0.693 & 0.709 & 0.662 & \textbf{0.811} & 0.732 & 0.745 & 0.424 \\
necrosis & 0.59 & 0.716 & \textbf{0.770} & 0.753 & 0.696 & 0.719 & 0.748 & 0.704 & 0.622 & 0.222 \\
mitotic activity & 0.83 & 0.778 &  & 0.778 & 0.750 & 0.833 & 0.583 & 0.500 & \textbf{0.917} & 0.083 \\
multifocal & 0.74 & 0.645 &  & 0.571 & 0.565 & 0.524 & 0.583 & 0.280 & \textbf{0.738} & 0.476 \\
infiltrate & 0.45 & 0.360 &  & 0.333 & 0.342 & 0.447 & 0.421 & \textbf{0.474} & 0.447 & 0.000 \\
precursor present & 0.93 & 0.888 & 0.896 & 0.904 & \textbf{0.919} & 0.887 & \textbf{0.919} & 0.500 & 0.100 & 0.125 \\
molecular subtype & 0.60 & 0.188 &  &  & 0.385 & 0.399 & \textbf{0.421} &  &  &  \\
tumour purity & 0.33 & 0.476 &  & \textbf{0.498} & 0.482 & 0.455 & 0.412 & 0.253 & 0.236 &  \\
leukocyte fraction & 0.36 & 0.435 &  & \textbf{0.460} & 0.448 & 0.441 & 0.398 & 0.201 & 0.152 &  \\
\midrule
high grade with LVI & 0.73 & 0.544 &  & 0.737 & 0.734 & \textbf{0.769} & \textbf{0.769} & 0.450 & 0.482 & 0.320 \\
necrosis with high mitotic & 1.00 & 0.500 &  & \textbf{1.000} & \textbf{1.000} & \textbf{1.000} & \textbf{1.000} & \textbf{1.000} & \textbf{1.000} & 0.000 \\
precursor type & 0.67 & 0.208 &  & 0.313 & 0.228 & 0.456 & 0.477 & 0.342 & \textbf{0.685} &  \\
\bottomrule
\end{tabular}
\end{table*}

%% file: tables/T4_router.tex
\begin{table*}[t]
\centering
\footnotesize\setlength{\tabcolsep}{3.5pt}
\caption{Routing between Opsis-S and Opsis-V and their probability fusion on the NuCLS test split (mean $\pm$ sd over 3 seed pairs). Opsis-S answers in 0.4 ms per question with the frozen encoder amortised over the questions of a field (0.1 ms for the decoder alone), Opsis-V in 8 ms. A question that a confidence or learned router escalates has already been answered by Opsis-S and is charged for both. The geometric mean of both is the most accurate and best calibrated answer.}
\label{tab:router}
\begin{tabular}{lrrrr}
\toprule
rule & accuracy $\uparrow$ & NLL $\downarrow$ & Opsis-V share & ms per question $\downarrow$ \\
\midrule
Opsis-S only & 0.791 $\pm$ 0.006 & 0.529 $\pm$ 0.028 & 0.0\% & 0.39 \\
Opsis-V only & 0.779 $\pm$ 0.003 & 0.563 $\pm$ 0.004 & 100.0\% & 8.10 \\
oracle & 0.861 $\pm$ 0.003 & 0.456 $\pm$ 0.019 & 7.0\% & 0.92 \\
per operator (chosen on val) & 0.792 $\pm$ 0.005 & 0.529 $\pm$ 0.023 & 14.1\% & 1.47 \\
confidence escalation, 10\% of val & 0.795 $\pm$ 0.003 & 0.539 $\pm$ 0.022 & 10.1\% & 1.21 \\
confidence escalation, 25\% of val & 0.799 $\pm$ 0.001 & 0.544 $\pm$ 0.015 & 24.6\% & 2.38 \\
confidence escalation, 50\% of val & 0.791 $\pm$ 0.003 & 0.550 $\pm$ 0.007 & 49.4\% & 4.39 \\
per operator + confidence (10\% of the rest) & 0.795 $\pm$ 0.002 & 0.537 $\pm$ 0.017 & 22.5\% & 2.16 \\
per operator + confidence (25\% of the rest) & 0.798 $\pm$ 0.002 & 0.542 $\pm$ 0.011 & 34.7\% & 3.14 \\
learned router, 10\% escalated & 0.795 $\pm$ 0.003 & 0.539 $\pm$ 0.023 & 10.2\% & 1.21 \\
learned router, 25\% escalated & 0.800 $\pm$ 0.001 & 0.542 $\pm$ 0.016 & 24.5\% & 2.37 \\
fusion: arithmetic mean of Opsis-S and Opsis-V & 0.805 $\pm$ 0.003 & 0.488 $\pm$ 0.008 & 100.0\% & 8.49 \\
fusion: geometric mean of Opsis-S and Opsis-V & 0.805 $\pm$ 0.002 & 0.487 $\pm$ 0.007 & 100.0\% & 8.49 \\
\bottomrule
\end{tabular}
\end{table*}

%% file: tables/T7_ways.tex
\begin{table*}[t]
\centering
\footnotesize\setlength{\tabcolsep}{3pt}
\caption{Three ways for a frontier model to read a slide, on the 189 breast questions (majority class 0.667). The model asks typed questions of Opsis-S, looks at pixels, or reads a written description and answers from the text alone. Asking minus this is the paired difference in points with its 95 percent bootstrap interval over slides. Tokens in is the mean input per question as each interface reports it, regions the number of regions the model chose to view per question, words the length of the description, and to write the tokens spent writing it, paid once per slide. In the grey image row every image of the row above is replaced by a grey image of the same size.}
\label{tab:ways}
\begin{tabular}{lrrrrrr}
\toprule
condition & acc. & asking minus this & tokens in & regions & words & to write \\
\midrule
\multicolumn{7}{l}{\textit{Claude Opus 5.5}} \\
asks Opsis-S & 0.794 &  & 6,732 &  &  &  \\
looks at a thumbnail and six tiles & 0.656 & +13.8 [4.6, 23.2] & 50,374 &  &  &  \\
looks at views of its own choice & 0.608 & +18.5 [8.5, 29.5] & 25,638 & 5.4 &  &  \\
reads the PRISM2 report & 0.630 & +16.4 [5.9, 27.8] & 2,038 &  & 78 &  \\
reads its own description of 6 tiles & 0.582 & +21.2 [11.0, 31.4] & 4,587 &  & 927 & 58,356 \\
reads its own description of 48 tiles & 0.577 & +21.7 [9.0, 34.7] & 16,372 &  & 6,140 & 350,682 \\
\addlinespace[2pt]\multicolumn{7}{l}{\textit{gpt-6-sol}} \\
asks Opsis-S & 0.788 &  & 1,448 &  &  &  \\
looks at a thumbnail and six tiles & 0.582 & +20.6 [11.0, 29.8] & 2,995 &  &  &  \\
looks at grey images in their place & 0.460 & +32.8 [20.7, 44.8] & 2,995 &  &  &  \\
looks at views of its own choice & 0.587 & +20.1 [9.1, 31.6] & 97,186 & 10.1 &  &  \\
reads the PRISM2 report & 0.630 & +15.9 [5.0, 27.4] & 240 &  & 78 &  \\
reads the description of 6 tiles by Claude Opus 5.5 & 0.503 & +28.6 [17.4, 39.8] & 1,415 &  & 927 & 58,356 \\
reads the description of 48 tiles by Claude Opus 5.5 & 0.534 & +25.4 [13.4, 38.2] & 8,718 &  & 6,140 & 350,682 \\
\addlinespace[2pt]\multicolumn{7}{l}{\textit{gpt-5.5}} \\
asks Opsis-S & 0.783 &  & 1,711 &  &  &  \\
looks at a thumbnail and six tiles & 0.582 & +20.1 [7.9, 31.7] & 2,995 &  &  &  \\
\addlinespace[2pt]\multicolumn{7}{l}{\textit{gpt-6-astra}} \\
asks Opsis-S & 0.788 &  & 1,417 &  &  &  \\
looks at a thumbnail and six tiles & 0.577 & +21.2 [10.1, 31.8] & 2,995 &  &  &  \\
\bottomrule
\end{tabular}
\end{table*}

%% file: tables/T8_ways_pan.tex
\begin{table*}[t]
\centering
\footnotesize\setlength{\tabcolsep}{3pt}
\caption{Asking against reading on the whole pan cancer test split: the seven report fields and five composites of \cref{tab:ways} on every test slide with a reported field (3,797 questions on 1,179 slides). Claude Opus 5.5 and gpt-6-sol ask typed questions of Opsis-S, and gpt-6-sol reads the report PRISM2 writes for the slide and answers from the text alone. Asks minus reads is the paired difference of the two gpt-6-sol conditions in points with its 95 percent bootstrap interval over slides. The two models give the same answer on 96.8 percent of the questions when asking. The mean input per question is 1,329 tokens for gpt-6-sol asking, 208 reading and 6,620 for Claude Opus 5.5 through the Claude Code CLI. Bold is the best method of a row.}
\label{tab:ways_pan}
\begin{tabular}{lrrrrrr}
\toprule
question & n & maj. & Opus 5.5 asks & gpt-6-sol asks & gpt-6-sol reads & asks minus reads \\
\midrule
\multicolumn{7}{l}{\textit{Report fields}} \\
grade & 764 & 0.45 & \textbf{0.682} & 0.660 & 0.579 & +8.1 [4.3, 11.9] \\
LVI & 754 & 0.60 & \textbf{0.666} & 0.659 & 0.621 & +3.8 [0.4, 7.3] \\
PNI & 186 & 0.53 & \textbf{0.769} & \textbf{0.769} & 0.699 & +7.0 [1.1, 12.9] \\
necrosis & 351 & 0.66 & \textbf{0.849} & 0.829 & 0.521 & +30.8 [23.6, 37.6] \\
precursor present & 277 & 0.81 & \textbf{0.834} & 0.827 & 0.592 & +23.5 [16.2, 30.7] \\
mitotic activity & 217 & 0.61 & 0.742 & 0.747 & \textbf{0.765} & -1.8 [-8.3, 4.1] \\
multifocal & 253 & 0.56 & \textbf{0.664} & \textbf{0.664} & 0.585 & +7.9 [1.6, 14.2] \\
\addlinespace[2pt]\multicolumn{7}{l}{\textit{Composites}} \\
high grade with LVI & 479 & 0.76 & \textbf{0.797} & 0.779 & 0.752 & +2.7 [-0.8, 6.3] \\
precursor type & 222 & 0.50 & 0.685 & \textbf{0.694} & 0.635 & +5.9 [1.4, 10.8] \\
necrosis with high mitotic & 146 & 0.71 & 0.822 & 0.815 & \textbf{0.829} & -1.4 [-10.3, 7.5] \\
grade from components & 98 & 0.51 & 0.673 & 0.663 & \textbf{0.755} & -9.2 [-18.4, 0.0] \\
DCIS with comedo necrosis & 50 & 0.74 & 0.660 & \textbf{0.760} & 0.420 & +34.0 [12.0, 54.0] \\
\midrule
all report fields & 2,802 & 0.58 & \textbf{0.722} & 0.712 & 0.607 & +10.5 [8.4, 12.5] \\
all composites & 995 & 0.67 & \textbf{0.757} & 0.753 & 0.721 & +3.2 [0.3, 6.1] \\
all questions & 3,797 & 0.60 & \textbf{0.731} & 0.722 & 0.637 & +8.6 [6.6, 10.5] \\
\bottomrule
\end{tabular}
\end{table*}

%% file: fig/figS_fov_radar.tex
\begin{figure}[t]
\centering
\includegraphics[width=\linewidth]{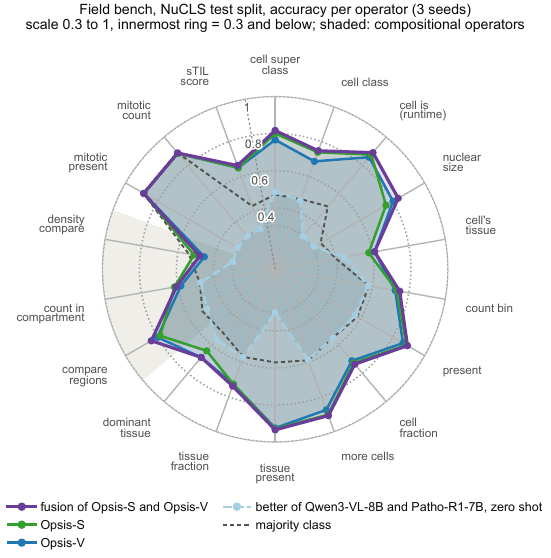}
\caption{Field bench on the NuCLS test split, accuracy per operator, mean over 3 seeds, one axis per operator (TIL location, with 12 test questions, is not drawn, the compositional operators are in the shaded sector, radial scale 0.3 at the innermost ring to 1 at the rim). \methS, \methV and their fusion against the better of two zero shot vision language models (Qwen3-VL-8B, Patho-R1-7B) answering the same typed questions from the same images, and the majority class.}
\label{fig:fov}
\end{figure}

%% file: fig/fig5_transfer.tex
\begin{figure*}[t]
\centering
\includegraphics[width=\textwidth]{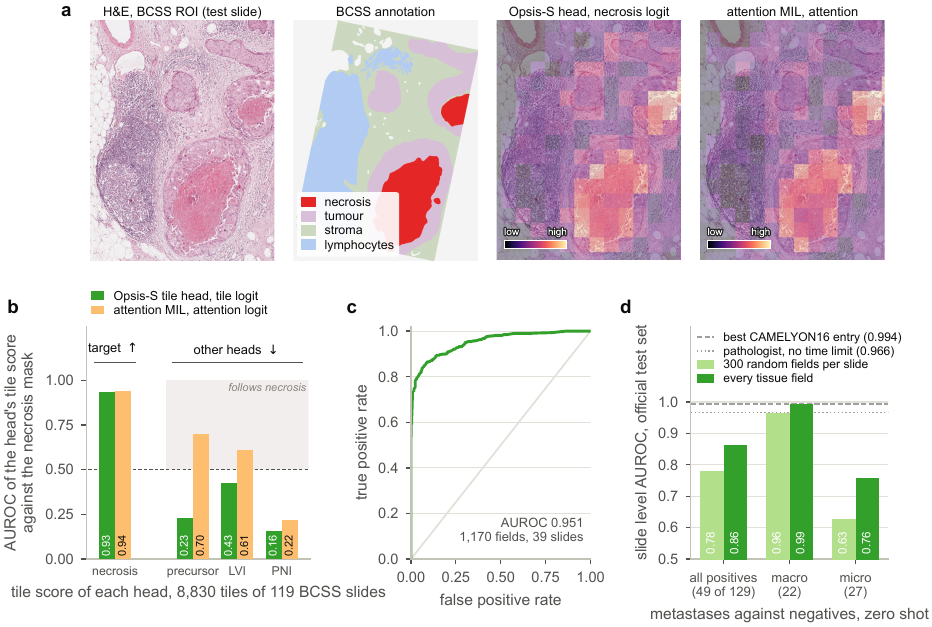}
\caption{\textbf{Localisation and transfer.} (a) A BCSS region of a held out TCGA-BRCA slide: H\&E, the BCSS annotation, the necrosis logit of the \methS tile head trained on report labels alone (noisy-OR pooling), and the attention map of the same tile head with gated attention pooling. (b) Specificity: the tile score of each head against the same necrosis mask (AUROC over 8,830 tiles of 119 BCSS slides). The necrosis head should follow the mask (target, higher is better), and the two necrosis maps are equally good. The other heads should not (lower is better): the attention of the precursor head also follows necrosis (0.70 against 0.23, in the shaded band). Attention marks where a model looks, the per operator logit marks what it finds. (c) CAMELYON16, never seen in training: the field level probability of tumour cells from the field questions of \methS against the lesion masks on 1,170 fields of 39 test slides, zero shot. (d) Slide level AUROC on the official 129 slide test set, with the maximum field probability over 300 random fields per slide or the top 5 mean over every tissue field on an 80 um grid, for all positives, macro metastases and micro metastases against negatives. The reference lines are AUCs over all positives from the challenge report (best entry 0.994, trained on the cohort, and a pathologist without time constraint 0.966) and compare with the all positives bars only.}
\label{fig:transfer_full}
\end{figure*}

%% file: fig/fig5b_camelyon_maps.tex
\begin{figure*}[t]
\centering
\includegraphics[width=\textwidth]{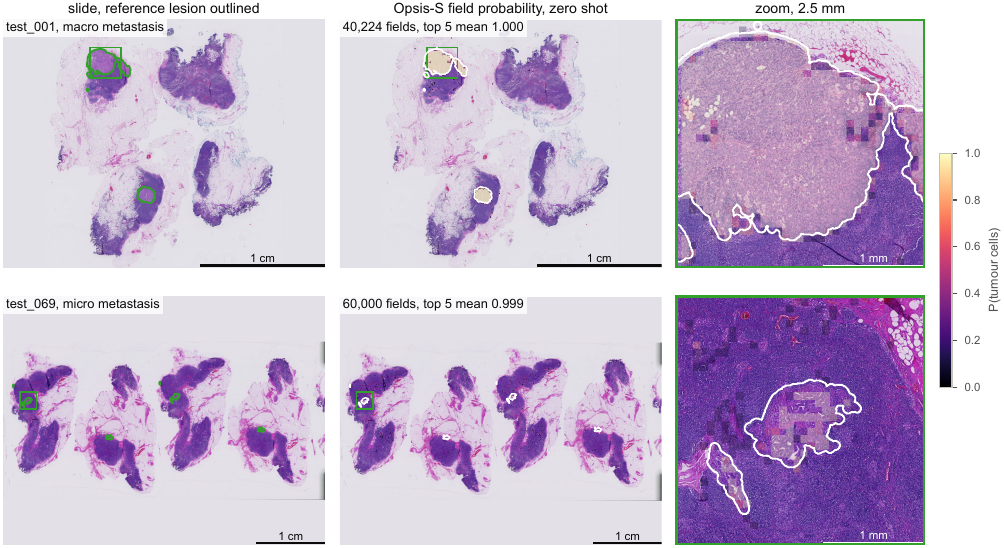}
\caption{Zero shot metastasis maps on two slides of the official CAMELYON16 test set, one macro and one micro metastasis, from the field decoder, which never saw a lymph node or a CAMELYON slide. Left: the slide at low power with the reference lesion outlined in green. Middle: the probability of tumour cells that \methS assigns to every 80 um tissue field on a non overlapping grid, drawn where it exceeds 0.05. Right: a 2.5 mm zoom on the largest lesion (green box) with the field probabilities over the H\&E and the reference outline in white. Scale bars 1 cm and 1 mm. The slide score is the mean of the five highest field probabilities.}
\label{fig:camelyon}
\end{figure*}

%% file: fig/fig3_organiser.tex
\begin{figure*}[t]
\centering
\includegraphics[width=\textwidth]{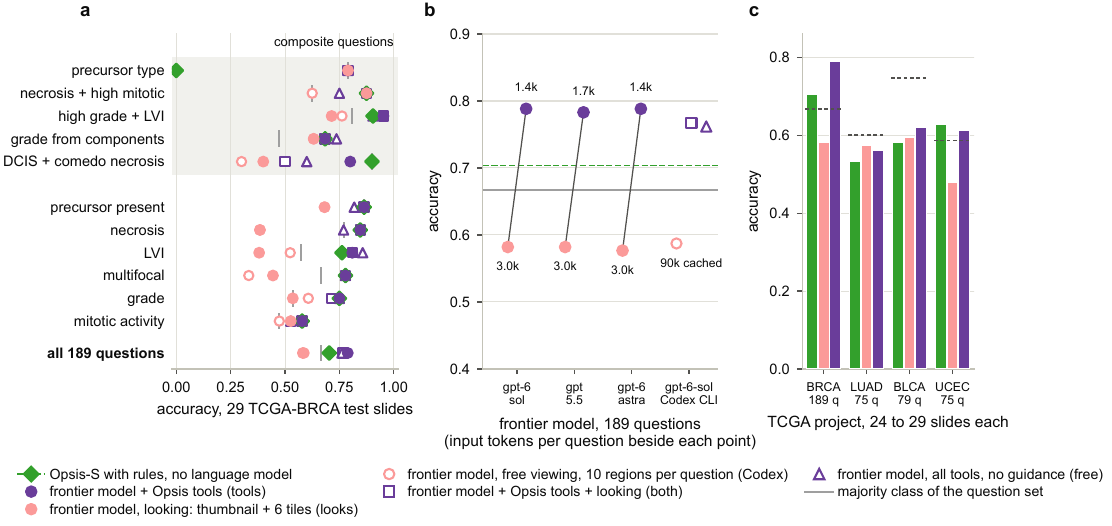}
\caption{\textbf{Organiser study.} (a) Accuracy per operator on 189 synoptic and composite questions about 29 TCGA-BRCA test slides for six conditions: \methS with fixed rules and no language model, which has a rule for four of the five composite questions (shaded) and counts as wrong on precursor type, gpt-6-sol with \method tools (tools), gpt-6-sol looking at a thumbnail and six prescribed 256 um tiles (looks), gpt-6-sol with free viewing through the Codex CLI (ten regions per question), looking plus tools (both), and all tools with no guidance (free). Grey ticks mark the majority class of each operator. (b) The three OpenAI models, looking against \method tools, with the input tokens per question beside each point. The fourth column is gpt-6-sol through the Codex CLI, whose cached system prompt costs 90k tokens per question. The dashed line is \methS with rules, the solid grey line the majority class. (c) The same three conditions on 24 to 29 test slides each of four TCGA projects (gpt-6-sol, through the API for BRCA and the Codex CLI for the others). Where the slide decoder is above the majority class (breast) the tools lead looking by 21 points. In lung, bladder and endometrium, where \methS with rules is below the majority class or four points above it (the dashed line), asking stays within four points of the rules: the organiser is only as good as the model it asks.}
\label{fig:organiser_dots}
\end{figure*}

%% file: fig/fig4_pan_comparables.tex
\begin{figure*}[t]
\centering
\includegraphics[width=\textwidth]{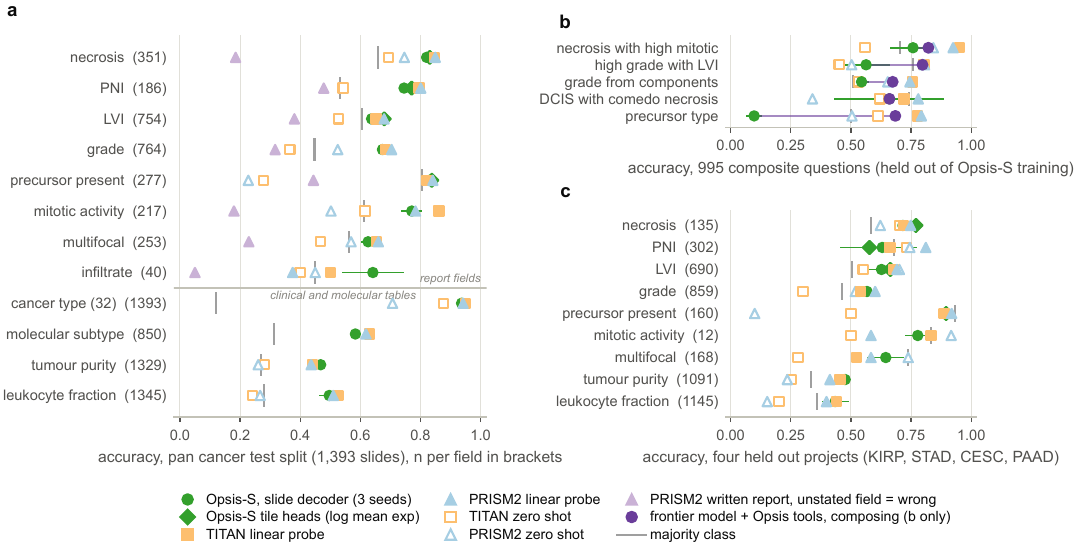}
\caption{Pan cancer slide level comparables on the pan cancer test split (1,393 slides of 28 projects). (a) Accuracy per field for \methS (the K = 4,096 slide decoder over 3 seeds, and the full bag tile heads) against slide foundation models with labels (TITAN and PRISM2 linear probes, one classifier per field, trained on our split) and without labels (TITAN and PRISM2 zero shot, and the report PRISM2 writes for the slide, extracted with the TCGA schema with an unstated field counted as wrong). Report fields are agreement with the pathology report, the lower block comes from the clinical and molecular tables. The grey tick is the majority class. With labels every method lands in the same place, since the state is the ceiling. Without labels only the typed questions answer every field. (b) The 995 composite questions, which \methS was never trained on: probes trained on them with labels, zero shot models, and the organiser (Claude Opus 5.5) composing \methS primitives. (c) Four projects held out from training (KIRP, STAD, CESC, PAAD).}
\label{fig:pan_dots}
\end{figure*}